\documentclass[10pt,twocolumn,letterpaper]{article}

\usepackage{float}
\usepackage[pagenumbers]{cvpr} % To force page numbers, e.g. for an arXiv version
\usepackage[dvipsnames]{xcolor}

\definecolor{cvprblue}{rgb}{0.21,0.49,0.74}
\usepackage[pagebackref,breaklinks,colorlinks,citecolor=cvprblue]{hyperref}
\usepackage[accsupp]{axessibility} % Improves PDF readability for those with visual impairments.
\def\paperID{10192} % *** Enter the Paper ID here
\def\confName{CVPR}
\def\confYear{2024}

\title{HIPTrack: Visual Tracking with Historical Prompts}
\author{Wenrui Cai$^1$, Qingjie Liu$^{1,2,3,*}$, Yunhong Wang$^{1,3}$\\
$^1$State Key Laboratory of Virtual Reality Technology and Systems, Beihang University, Beijing, China  \\
$^2$Zhongguancun Laboratory, Beijing, China  \\
$^3$Hangzhou Innovation Institute, Beihang University, Hangzhou, China \\
{\tt\small \{wenrui\_cai, qingjie.liu, yhwang\}@buaa.edu.cn}
}

\begin{document}
\maketitle
\begin{abstract}
Trackers that follow Siamese paradigm utilize similarity matching between template and search region features for tracking. 
Many methods have been explored to enhance tracking performance by incorporating tracking history to better handle scenarios involving target appearance variations such as deformation and occlusion. 
However, the utilization of historical information in existing methods is insufficient and incomprehensive, which typically requires repetitive training and introduces a large amount of computation.
%Many efforts have been made to enhance the performance of trackers in scenarios involving target appearance variations by incorporating tracking history context. Many methods have attempted to enhance tracker performance in scenarios involving target appearance variations, such as deformation and occlusion, by incorporating tracking history information.
%The key to improve the tracker to deal with the appearance change scenes such as target deformation and occlusion is to introduce historical context information.
%However, only using the template to predict the target in each subsequent frame will not be able to effectively deal with the situations such as deformation, occlusion and other appearance changes. 
%However, the existing trackers either use limited and imprecise historical information or introduce a large amount of computation when introducing historical context information. 
In this paper, we show that by providing a tracker that follows Siamese paradigm with precise and updated historical information, a significant performance improvement can be achieved with completely unchanged parameters.
Based on this, we propose a \textbf{hi}storical \textbf{p}rompt network that uses refined historical foreground masks and historical visual features of the target to provide comprehensive and precise prompts for the tracker.
We build a novel tracker called \textbf{HIPTrack} based on the historical prompt network, which achieves considerable performance improvements without the need to retrain the entire model.
%Most trackers perform template and search region similarity matching to find the most similar object to the template during tracking. 
%However, they struggle to make prediction when the target appearance changes due to the limited historical information introduced by roughly cropping the current search region based on the predicted result of previous frame.
%These trackers utilize limited historical information by coarsely cropping the current search region based on the predicted target state from the previous frame and encountering challenges in prediction when the target undergoes appearance changes. 
%In this paper, we identify that the central impediment to improving the performance of existing trackers is the incapacity to integrate abundant and effective historical information. To address this issue, we propose a Historical Information Prompter (HIP) to enhance the provision of historical information. We also build HIPTrack upon HIP module. HIP is a plug-and-play module that make full use of search region features to introduce historical appearance information. It also incorporates historical position information by constructing refined mask of the target. HIP is a lightweight module to generate historical information prompts. By integrating historical information prompts, HIPTrack significantly enhances the tracking performance without the need to retrain the backbone. 
We conduct experiments on seven datasets and experimental results demonstrate that our method surpasses the current state-of-the-art trackers on LaSOT, LaSOT$_{ext}$, GOT-10k and NfS. Furthermore, the historical prompt network can seamlessly integrate as a plug-and-play module into existing trackers, providing performance enhancements. The source code is available at \href{https://github.com/WenRuiCai/HIPTrack}{https://github.com/WenRuiCai/HIPTrack}.
\end{abstract}
\renewcommand{\thefootnote}{}
\footnote{$^*$Corresponding author.}
\section{Introduction}\label{sec:intro}
%Most trackers perform similarity matching between template and search region features to track. 
%However, only using the template to predict the target in each subsequent frame will not be able to effectively deal with the situations such as deformation, occlusion and other appearance changes. 
Visual tracking aims to estimate the position of a target in subsequent frames of a video, given the initial state in the first frame 
%Most trackers still follow the siamese paradigm for tracking 
\cite{bertinetto2016fully,xu2020siamfc++,li2019siamrpn++,chen2021transformer}.  
%Traditionally, most trackers define visual tracking as the problem of finding the target with the highest similarity to the template within a search region   \cite{bertinetto2016fully,xu2020siamfc++,li2019siamrpn++,chen2021transformer}. 
It can be challenging when the target experiences deformation, scale variation, and partial occlusion over time. While Transformer-based one-stream trackers \cite{Gao_2023_CVPR_GRM,chen2022backbone,ye_2022_joint,Cui_2022_MixFormer} have significantly improved the tracking accuracy by incorporating feature extraction and interaction of template and search regions into a single Transformer, they still follow Siamese paradigm that performs similarity matching between template and search regions. Only using the template to predict the target in subsequent frames may not be effective in handling complex scenarios.

\begin{figure}[t]
%是可选项 h表示的是here在这里插入，t表示的是在页面的顶部插入
\centering
\includegraphics[scale=0.10]{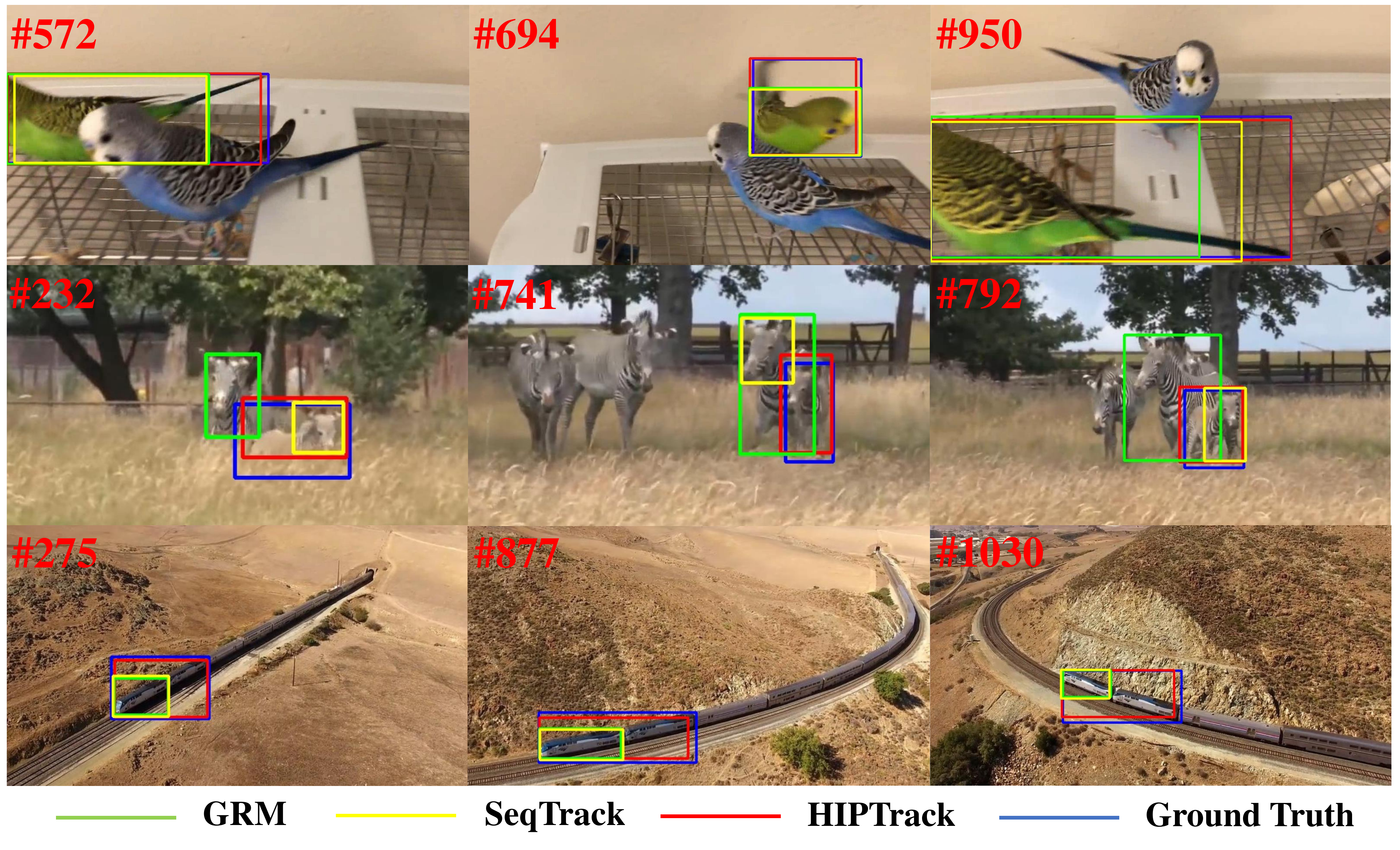}
\vspace{-4.5ex}
\caption{Visualized comparisons of our approach and other excellent trackers GRM \cite{Gao_2023_CVPR_GRM} and SeqTrack   \cite{Chen_2023_CVPR_seqtrack}. Our method performs better when the target suffer from occlusion, deformation and scale variation.}
%\vspace{-2ex}
\label{fig:1}
\end{figure}

There have been numerous attempts to incorporate historical context into trackers using techniques such as space-time memory networks \cite{fu2021stmtrack}, utilizing a tracked frame as auxiliary template \cite{Cui_2022_MixFormer,yan2021learning} and incorporating multiple frames as inputs to the backbone  \cite{wang2021transformer,he2023target_TATrack}.
%Many efforts have been made to incorporate historical context into trackers by employing space-time memory networks \cite{fu2021stmtrack}, utilizing a tracked frame as auxiliary template \cite{Cui_2022_MixFormer,yan2021learning} or incorporating multiple frames as inputs to the backbone  \cite{wang2021transformer,he2023target_TATrack}. 
However, the target position 
 introduced by space-time memory networks lacks accuracy, while the use of two backbones for the current frame and tracked frames introduces a substantial number of parameters. Utilizing a tracked frame as auxiliary template is insufficient and susceptible to introducing distractors. Employing multiple frames as inputs introduces considerable computational complexity while inadequately preserving a substantial amount of historical context. Recently, autoregressive trackers \cite{Wei_2023_CVPR_autoregressive,Chen_2023_CVPR_seqtrack} have achieved state-of-the-art performance by tokenizing the target position and incorporating a generative decoder to make predictions. Autoregressive trackers can make use of multi-frame historical position information for prediction, and auxiliary template is also introduced in \cite{Chen_2023_CVPR_seqtrack}. However, autoregressive trackers still have not fully utilized the historical visual features and require full parameter training with a substantial parameter burden.

In this paper, we conduct an analysis on several trackers \cite{ye_2022_joint,Cui_2022_MixFormer,chen2022backbone,chen2021transformer} that follow Siamese paradigm. We provide updated templates and more accurately cropped search regions while keeping the model unchanged. The results in Figure \ref{fig:analysis} show significant performance improvements, indicating that high quality historical prompt can improve tracking accuracy without the need for a full parameter retraining. 
%In both cases, significant performance improvements can be observed, which implies that we only need to provide high quality historical prompts for trackers to improve tracking accuracy, without the need for a full parameter retraining. 
Based on this analysis, we propose \textbf{HIPTrack}, a novel tracker that features the \textbf{hi}storical \textbf{p}rompt network as its core module. 
This lightweight module includes an encoder for encoding historical target feature with both positions and visual features of the target, and a decoder for generating  historical prompt for current search region.
Compared with \cite{fu2021stmtrack}, HIPTrack incorporates more precise target position information without requiring an additional backbone.
The historical prompt encoder introduces precise target position by constructing refined foreground masks, which are then encoded along with the visual features of the target as the historical target features.
Compared with \cite{yan2021learning,Cui_2022_MixFormer,Wei_2023_CVPR_autoregressive,Chen_2023_CVPR_seqtrack}, HIPTrack introduces more sufficient and comprehensive historical information.
We store a large amount of historical target features in the historical prompt decoder and generate prompts tailored to the current search region through adaptive decoding. By leveraging the concept of prompt learning to introduce historical information, 
HIPTrack achieves significant performance improvements while still maintaining tracking efficiency.
Additionally, only the historical prompt network and prediction head need to be trained in HIPTrack, reducing the number of training parameters by more than 80\% compared with \cite{Wei_2023_CVPR_autoregressive}.

We evaluate the performance of our method on 7 datasets and
experimental results show that HIPTrack achieves state-of-the-art performance on LaSOT, LaSOT$_{\mathrm{ext}}$, GOT-10k and NfS. Figure \ref{fig:1} clearly demonstrates the improved capability of our method in handling scenarios that involve occlusion, deformation and scale variation. The main contributions of this work can be summarized as:  \textbf{1)} We propose the historical prompt network, a module that encode high quality historical target features and generate effective historical prompts for tracking.  \textbf{2)} We propose a novel tracker called HIPTrack based on the historical prompt network that eliminates the need for retraining the entire model. \textbf{3)} Experimental results show that HIPTrack outperforms all trackers on LaSOT, LaSOT$_{\mathrm{ext}}$, GOT-10k, and NfS; Additional experimental results demonstrate that the historical prompt network can serve as a plug-and-play component to improve the performance of current trackers.

\section{Related Work}
\label{sec:related}

\subsection{Trackers without Historical Information}

Trackers that follow Siamese paradigm perform similarity matching between template and search region
%use a template and search region similarity matching method
for tracking   \cite{bertinetto2016fully,li2019siamrpn++,zhang2020ocean,xu2020siamfc++}, which do not leverage historical information. These trackers rely on the predicted bounding box from the previous frame to crop the current search region. 
Recently, several trackers have employed Transformers to enhance the template-search region interaction, such as TransT  \cite{chen2021transformer}, DTT \cite{yu2021high} and SparseTT
  \cite{fu2022sparsett}. One-stream trackers such as OSTrack \cite{ye_2022_joint}, SimTrack \cite{chen2022backbone}, GRM \cite{Gao_2023_CVPR_GRM} and DropTrack  \cite{Wu_2023_CVPR_dropmae} integrate feature extraction and interaction into one Transformer and significantly boost the tracking performance. However, these methods still do not incorporate any historical information and underperform in complex scenarios such deformation, scale variation and partial occlusion.

\subsection{Trackers with Historical Information}
STMTrack \cite{fu2021stmtrack} utilizes space-time memory networks \cite{oh2019video} to integrate historical information, but its non-shared backbones result in an excessive number of parameters. The template-free design may lead to the neglect of essential information and the target position description is not precise. 
STARK  \cite{yan2021learning} and MixFormer  \cite{Cui_2022_MixFormer} utilize one search region with high score as an auxiliary template, which is vulnerable to distractors.
%and reduces reliability. 
TATrack \cite{he2023target_TATrack} and TrDiMP \cite{wang2021transformer} design backbones capable of processing multiple frames to extract historical features, but the computational overhead makes it challenging to retain more historical frames. ARTrack \cite{Wei_2023_CVPR_autoregressive} and SeqTrack  \cite{Chen_2023_CVPR_seqtrack} utilize predicted coordinates of bounding box and employ a generative decoder to integrate historical positions across multiple frames with the current search region feature and make prediction. However, although SeqTrack \cite{Chen_2023_CVPR_seqtrack}  introduces an auxiliary template, both methods still have not fully exploited the historical visual features of the target. Additionally, autoregressive trackers require full parameter training, ARTrack \cite{Wei_2023_CVPR_autoregressive} even requires two-stage training, leading to significant training resource overhead. 

\subsection{Prompt Learning}
%这里再考虑一下
Prompt learning is a technique used to customize pre-trained models for specific tasks, which has been applied in the fields such as computer vision  \cite{jia2022visualPromptTuning,wang2022learningtopromptforcontinual}, natural language processing   \cite{lester2021power,li2021prefix}, and multimodal studies  \cite{zhou2022learning_to_prompt,Khattak_2023_CVPR_MaPLe,pmlr-v139-radford21a,Rao_2022_CVPR_DenseCLIP,Zhu_2023_CVPR_visual_prompt_multimodal_tracking}.
Prompt learning typically adjusts the model input or utilizes adapters at various layers to modify the input-output space. 
In the field of visual tracking, trackers that follow Siamese paradigm can be conceptualized as pre-trained models for feature similarity matching. By incorporating historical information prompts, the similarity matching ablity of these trackers can be extended to the temporal dimension.

\section{Method}
\label{sec:method}

\subsection{Revisit Current Trackers}

\begin{figure}[!t] \centering    
\subfigure[] {
 \label{fig_analysis:a}     
\includegraphics[width=0.46\columnwidth]{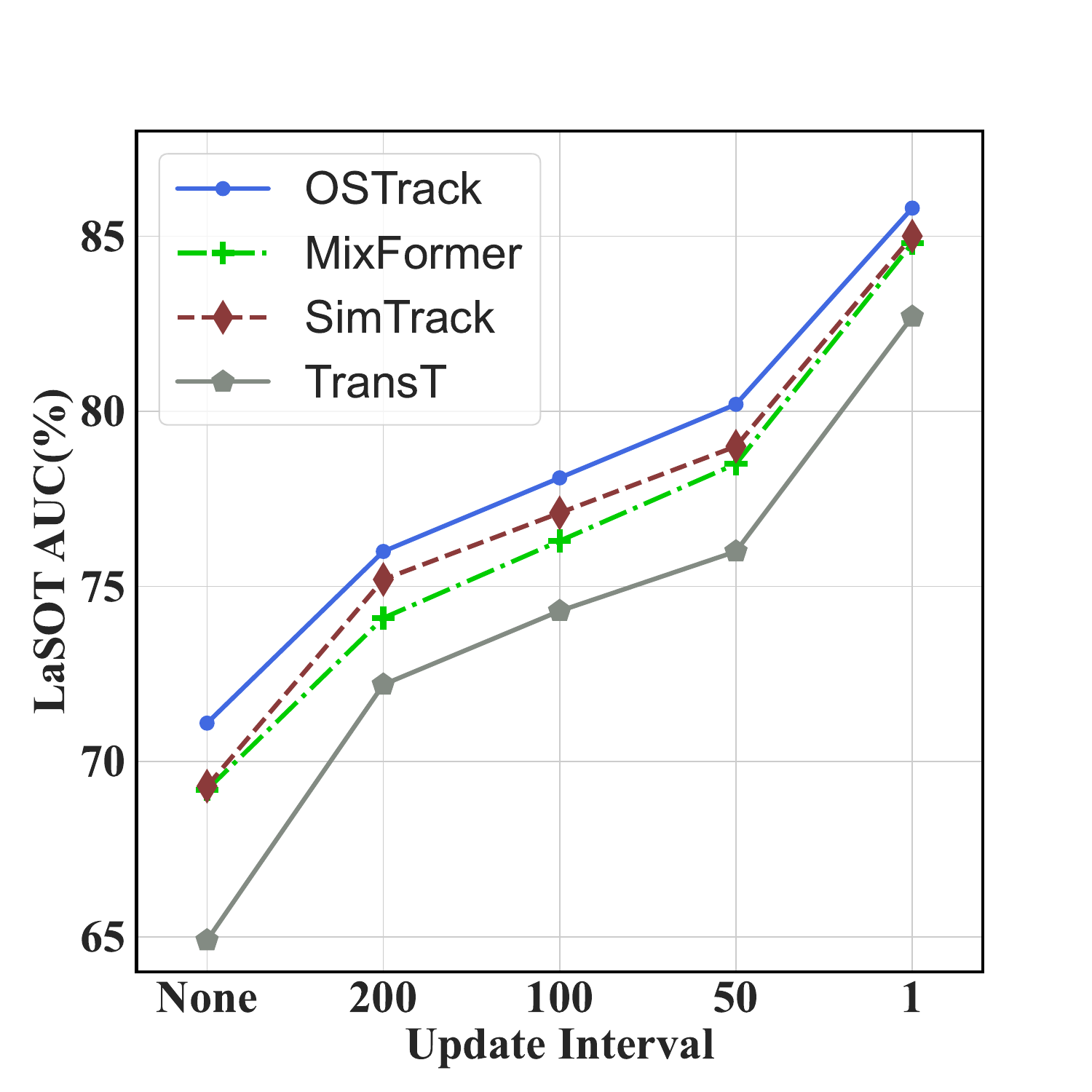}  
}     
\subfigure[] { 
\label{fig_analysis:b}     
\includegraphics[width=0.46\columnwidth]{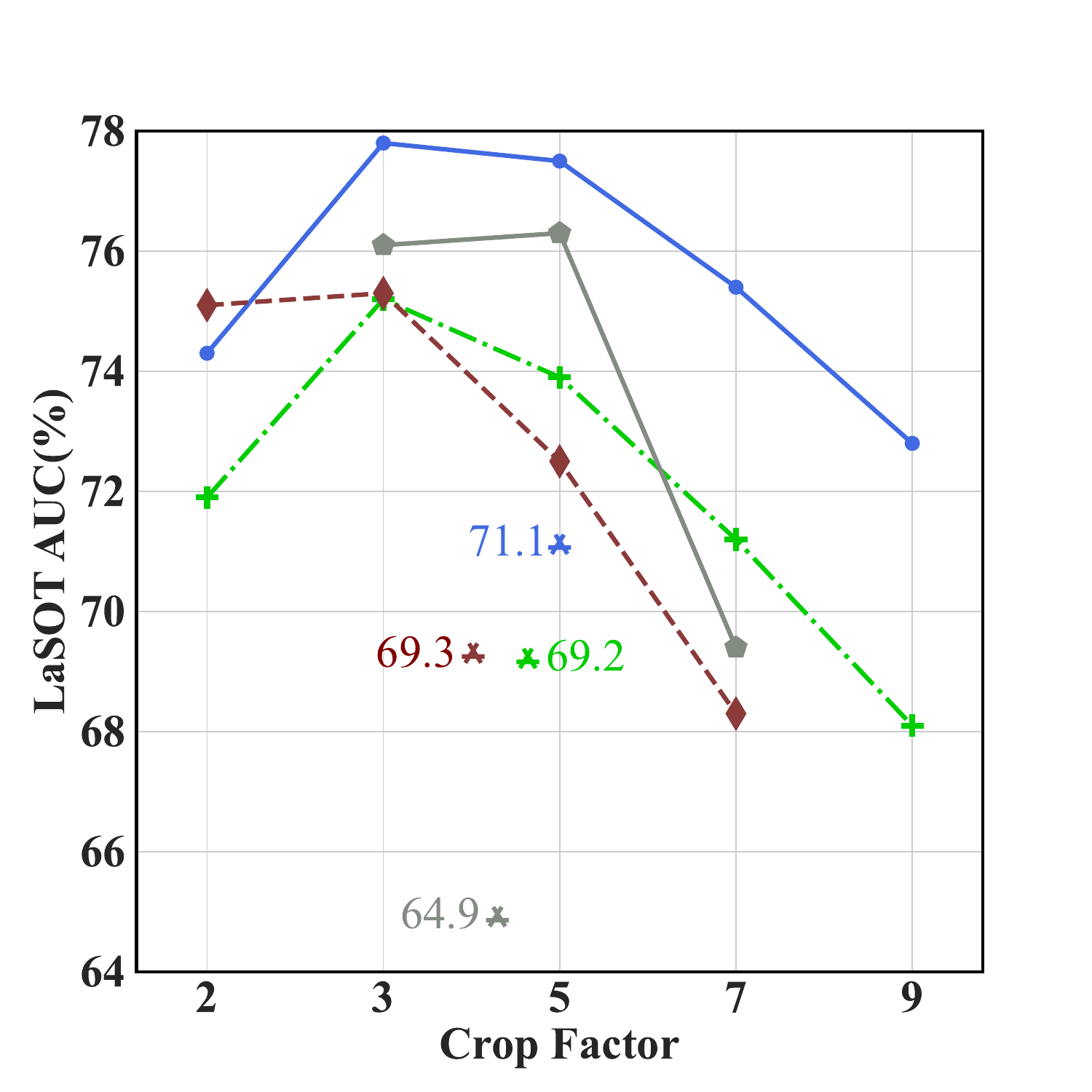}     
}    
\vspace{-2ex}
\caption{(a) shows the varying performance of trackers on LaSOT \cite{fan2019lasot} as the template update intervals change. (b) shows the varying performance of trackers on LaSOT as the crop factor of current search regions change. A larger crop factor indicates coarser cropping. Each cross symbol represents the baseline of the corresponding color method. Note that TransT \cite{chen2021transformer} does not have a fixed crop factor, so we choose to use an average crop factor instead. }     
\label{fig:analysis}

\end{figure}

%\begin{figure}[h]
%\centering
%{
%\begin{minipage}{4.0cm}
%\centering
%\includegraphics[scale=0.28]{figures/zhexian1.pdf} %以这幅图的0.5倍大小输出
%\end{minipage}
%}
%{
%\begin{minipage}{4.0cm}
%\centering
%\includegraphics[scale=0.28]%{figures/zhexian2.pdf}%以这幅图的0.5倍大小输出
%\end{minipage}
%}
%\caption{This figure shows the comparison of SparseTT with other trackers in the Precision plot on OTB2015, which includes eleven special scenarios such as Low Resolution, Motion Blur, Scale Variation, etc., and also includes the overall results of the dataset.}
%\label{fig:analysis}
%\end{figure}
To explore the performance ceiling of current trackers that follow Siamese paradigm, we conduct two experiments 
 on OSTrack  \cite{ye_2022_joint}, MixFormer \cite{Cui_2022_MixFormer}, SimTrack \cite{chen2022backbone} and TransT \cite{chen2021transformer}.
Firstly, we update the template every $\bm{n}$ frames, which is cropped with ground truth boxes. Despite a decrease in tracker performance with larger update intervals, as shown in Figure \ref{fig_analysis:a}, 
%it still outperforms the result that only use the first frame as template. 
when updating the template even with an interval as large as 200, the performance of trackers still remain superior to using only the initial frame as template, which means using more updated target visual features can significantly improve tracking performance. 
Secondly, we crop the search regions using ground truth boxes in the previous frame instead of the predicted boxes. Although the performance decreases with larger cropping factors, as shown in Figure \ref{fig_analysis:b}, when the cropping factor does not differ much from the baseline, the performance still significantly exceeds the baseline that uses predicted results in the previous frame to crop search regions, which suggests that providing the tracker with more accurate target location information can significantly improve performance.
Therefore, for current trackers that follow Siamese paradigm, a substantial performance boost can be attained by using effective historical target feature as prompts. 

\subsection{Overall Architecture}

\begin{figure*}[!t]
%是可选项 h表示的是here在这里插入，t表示的是在页面的顶部插入
\centering
\includegraphics[scale=0.6]{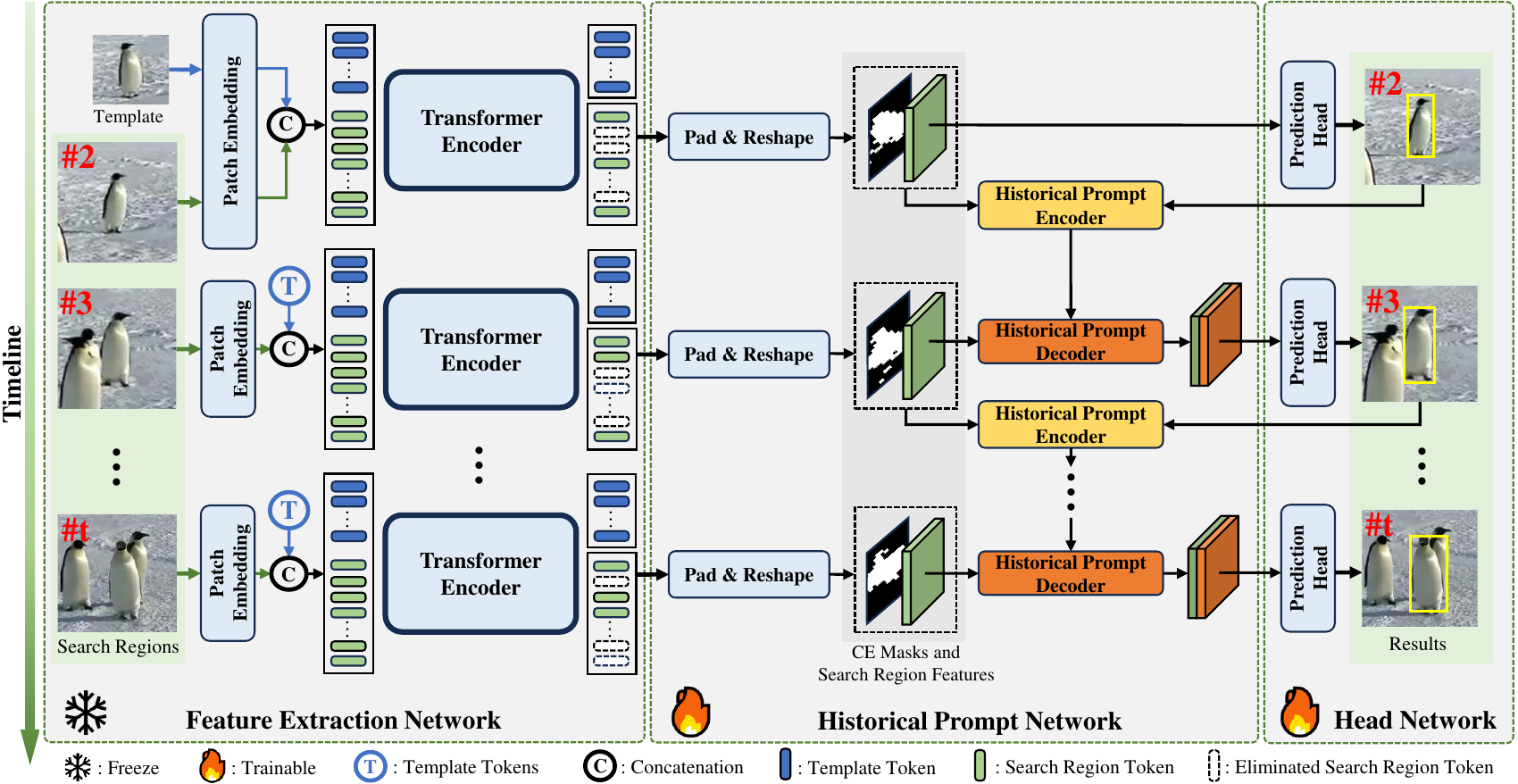}
\vspace{-2ex}
\caption{Overview of our proposed HIPTrack. The whole structure consists of a feature extraction network, a history prompt network, and a head prediction network. The historical prompt network comprises a historical prompt encoder and a historical prompt decoder. 
%The encoder is to leverage predictions from the current frame to encode historical prompts for subsequent frames, while the decoder generates the historical information prompts required for the current search region.
%The core of our method is (b) Historical Information Prompter, which consists of a lightweight convolutional network $\Phi$, a (c) Fusion Block that encode historical information prompt values and a (d) Historical Information Querier that gathers historical information prompt values from the memory bank and generate the historical information prompt for current search region. 
}
\label{fig:pipeline}
\end{figure*}

As shown in Figure  \ref{fig:pipeline}, we present HIPTrack, which comprises three main components: a feature extraction network,
%that extracts search region features interacted with template features
a historical prompt network, and a prediction head network. 
%The encoder is implemented using the Vision Transformer (ViT)  \cite{dosovitskiy2020image}.
The feature extraction network extracts the features of the search region interacted with the template, while filtering out background image patches of the search region.
%The feature extraction network employs a one-stream Vision Transformer (ViT) \cite{dosovitskiy2020image} architecture
%, wherein the template token is concatenated with the search region token and fed into the Transformer encoder. The Transformer extracts search region features interacted with template features. 
%The ViT is initialized from the weights of existing trackers, with its parameters fully frozen. 
The historical prompt network employs a historical prompt encoder to encode the position information and the visual features of the target from the current frame as historical target feature, and appends it to the memory bank in historical prompt decoder. 
%The memory bank increases in capacity as the number of tracking frames increases. 
In subsequent tracking, the historical prompt decoder generates historical prompt for each search region and concatenates historical prompt with compressed search region feature along the channel dimension. We adopt the same prediction head structure as OSTrack \cite{ye_2022_joint}. Due to the increase in the number of channels after incorporating the historical prompt, we introduce a residual convolutional layer at the input of the prediction head to reduce the channels. 

\subsection{Feature Extraction Network}

The feature extraction network utilizes a Vision Transformer (ViT) \cite{dosovitskiy2020image} as the visual backbone, initialized with the weights of existing trackers and its parameters are entirely frozen. The feature extraction process employs a one-stream approach. Within ViT, similar to OSTrack \cite{ye_2022_joint}, we incorporate an early Candidate Elimination (CE) module with the same elimination ratio. The CE module is embedded within the attention layers of ViT and filters out search region background tokens with the lowest attention scores in relation to the template token. Our method utilizes the image patches filtered by CE module to construct CE mask as one input of the historical prompt encoder.

\subsection{Historical Prompt Network}

%In the historical prompt encoder, the predicted bounding boxes are utilized to generate masks, which are then refined by removing the background area with Candidate Elimination (CE) module \cite{ye_2022_joint} in Transformer-based backbone to obtain more accurate historical foreground masks of the target.  
%The encoder encodes the historical target features by using both foreground masks that contain target position information and search region features that contain target visual feature in tracked frames. 
%In the historical prompt decoder, we store the encoded historical target feature along with its corresponding search region feature as an index in the memory bank, using a key-value pair structure. In each subsequent frame, the decoder uses the search region feature as a query and employs attention based on the Euclidean distance for adaptive aggregation of the historical target features to generate historical prompt for the current frame. This approach ensures that the queries and keys are in the same feature space and no other prediction results with possible errors are introduced.
The historical prompt network is the core module of our method, which consists of an encoder and a decoder, as shown in Figure \ref{fig:prompt}. 
Inspired by video object segmentation methods \cite{oh2019video,cheng2021rethinking}, we choose to use masks to describe the target position information. The predicted bounding boxes are utilized to generate masks, which are then refined by CE masks to obtain more accurate historical foreground masks of the target.
The encoder encodes the refined foreground masks and the target visual features together into the historical target features that serves as historical prompt value and appends it to the memory bank in the decoder, with the compressed search region feature as the key. The decoder utilizes the current search region feature a query and employs attention based on the Euclidean distance for adaptive aggregation of the historical target features to generate historical prompt for the current frame.

%The encoder employs the tracking result, search region feature and CE mask from the tracked frame to encode historical target feature that serves as historical prompt value and appends it to the memory bank in decoder, with the compressed search region feature as the key. 
%The decoder utilizes the search region feature to generate historical prompt suitable for the current frame.
\subsubsection{Historical Prompt Encoder}
%Historical prompt encoder employs a refined foreground mask to introduce accurate historical target position and utilizes search region feature to incorporate historical visual feature of the target. 
As shown in Figure \ref{fig:prompt}, we use the eliminated patches from the feature extraction network to construct a CE mask and generate a bounding box mask using the predicted bounding box. The two masks are combined using a bitwise $\mathrm{AND}$ operation to obtain the final refined foreground mask. The refined foreground mask, search region image, and search region feature are collectively utilized to encode the historical target feature, also referred to as prompt value.

%HIP provides effective historical information prompts for a tracker.
%It takes in a search region image from a particular frame, along with the search region features that interact with the template and the refined mask to generate historical information prompt values that are added to the memory bank.  
%Afterwards, for each search frame, the search region features go through the Historical Information Querier (HIQ) to retrieve historical information prompt values. HIQ aggregates all historical information prompt values to create customized historical prompt relevant to the current search region.
 Initially, the historical prompt encoder concatenates the input search region image with foreground mask along the channel dimension to create a new 4-channel image and feeds the image into a lightweight encoder represented as $\Phi$, yielding a feature map $\bm{K} \in  \mathbb{R}^{\frac{H}{16}\times \frac{W}{16} \times C_K}$ that contains target position information.  In our approach, $\Phi$ utilizes the first three stages of ResNet-18  \cite{he2016deep} to ensure that the downsampling factor aligns with that of the ViT backbone. Besides, we modify the input channel of the first convolutional layer of $\Phi$ to be 4 while keeping other channels unchanged. 

%Ultimately, HIP employs a fusion block to merge the feature map with the search region features that have already interacted with the template, resulting in the final encoded prompt value. To summarize, the entire process can be described as follows:
%\begin{equation}\label{eq_2}\small
%\begin{aligned}
% \bm{K} & = \Phi(\mathrm{Concat}(\bm{I}, \bm{M})) \\
 %\bm{P} & = \mathrm{Fusion}(\bm{K}, \bm{F}) \\
%\end{aligned}
%\end{equation}
%\noindent  where $\bm{I} \in \mathbb{R}^{H\times W\times 3}$ represents the search region image, $\bm{M} \in  \mathbb{R}^{H\times W\times 1}$ represents the mask of the target in the search region, $\bm{K} \in  \mathbb{R}^{\frac{H}{16}\times \frac{W}{16} \times C_K}$ represents the feature map with target position information generated by $\Phi$.
 %$\bm{F} \in  \mathbb{R}^{\frac{H}{16}\times \frac{W}{16} \times C_F}$ represents the reshaped search region features that have interacted with the template, $\bm{P} \in  \mathbb{R}^{\frac{H}{16}\times \frac{W}{16} \times C_P}$ is the output of the fusion block.
After obtaining the feature $\bm{K}$ that contains the target position information, the historical prompt encoder fuses $\bm{K}$ with the search region feature $\bm{F} \in  \mathbb{R}^{\frac{H}{16}\times \frac{W}{16} \times C_F}$. We concatenate $\bm{K}$ and $\bm{F}$ along the channel dimension and feed the result into a convolutional block with residual connections, yielding a fused feature map $\bm{F^{\prime}_1} \in \mathbb{R}^{\frac{H}{16}\times\frac{W}{16}\times C_{P}} $.
%which can be formulated as:
%\begin{equation}\label{eq_3_1}\small
%    \begin{aligned}
%        \bm{F^{\prime}_1} &= f_{RB1}([\bm{K}; \bm{F}]) \\
%    \end{aligned}
%\end{equation}
We denote the residual block as $f_{RB1}$, which consists of two $3\times 3$ convolutional layers with a padding of 1 and a stride of 1, the first convolutional layer has an output channel count of $C_{P}$ and the second convolutional layer maintains the same input and output channels.
%$\bm{F^{\prime}_1} \in \mathbb{R}^{\frac{H}{16}\times\frac{W}{16}\times C_{P}} $ is the output of the first residual block. 
Later, we adopt a Convolutional Block Attention Module \cite{Woo_2018_ECCV_CBAM} for feature enhancement of $\bm{F^{\prime}_1}$, which can be formulated as follows:
\begin{equation}\label{eq_3}\small
\begin{aligned}
 W_{C} = \mathrm{MLP}(&\mathrm{Pool}^{hw}_{max}(\bm{F^{\prime}_1})) + \mathrm{MLP}(\mathrm{Pool}^{hw}_{avg}(\bm{F^{\prime}_1}))\\
 W_{S} = \mathrm{Conv}(&[\mathrm{Pool}^{c}_{max}(\bm{F^{\prime}_1}); \mathrm{Pool}^{c}_{avg}(\bm{F^{\prime}_1})])\\
 %\bm{F^{\prime}_2} & = \sigma(W_{C}) \cdot \bm{F^{\prime}_1} \\
 \bm{F^{\prime}_2} = (\sigma(&W_{C}) \otimes \bm{F^{\prime}_1}) \odot \sigma(W_{S}) 
 %\bm{P} & = f_{RB2}(\bm{F^{\prime}_2})
\end{aligned}
\end{equation}

\noindent where $\mathrm{Pool}^{hw}_{max}$ and $\mathrm{Pool}^{hw}_{avg}$ represent max pooling and average pooling operations applied along the spatial pixel dimension, respectively. $\mathrm{Pool}^{c}_{max}$ and $\mathrm{Pool}^{c}_{avg}$ represent max pooling and average pooling operations along the channel dimension. The $\mathrm{MLP}$ consists of two linear layers and does not alter the dimensions, and $W_{C} \in \mathbb{R}^{C_{P}\times 1}$ is obtained by separately passing two pooling results through the $\mathrm{MLP}$ and then adding the two outputs up. $\mathrm{Conv}$ is a convolutional layer with a kernel size of 7, padding of 3, and an output channel count of 1. $W_{S} \in \mathbb{R}^{\frac{H}{16}\times\frac{W}{16} \times 1}$ is the output of $\mathrm{Conv}$. 
$\sigma$ represents Sigmoid function, $\otimes$, $\odot$ represent channel-wise multiplication and pixel-wise multiplication, respectively. $\bm{F_2^{\prime}} \in \mathbb{R}^{\frac{H}{16}\times\frac{W}{16}\times C_{P}}$ is the result of applying the channel weight and spatial weight to $\bm{F_1^{\prime}}$. 
We further add $\bm{F_1^{\prime}}$ with $\bm{F_2^{\prime}}$ and feed the added result into another residual block $f_{RB2}$ to obtain the final output $\bm{P} \in \mathbb{R}^{\frac{H}{16}\times\frac{W}{16}\times C_{P}}$ as the encoded prompt value. $f_{RB2}$ maintains the same structural configuration as $f_{RB1}$, with the distinction that $f_{RB2}$ does not alter the number of channels.
%:
% \begin{equation}\label{eq_3_2}\small
%\begin{aligned}
% \bm{P} & = f_{RB2}(\bm{F^{\prime}_2})
%\end{aligned}
%\end{equation}

\subsubsection{Historical Prompt Decoder}

\begin{figure}[!t]
%是可选项 h表示的是here在这里插入，t表示的是在页面的顶部插入
\centering
\includegraphics[scale=0.37]{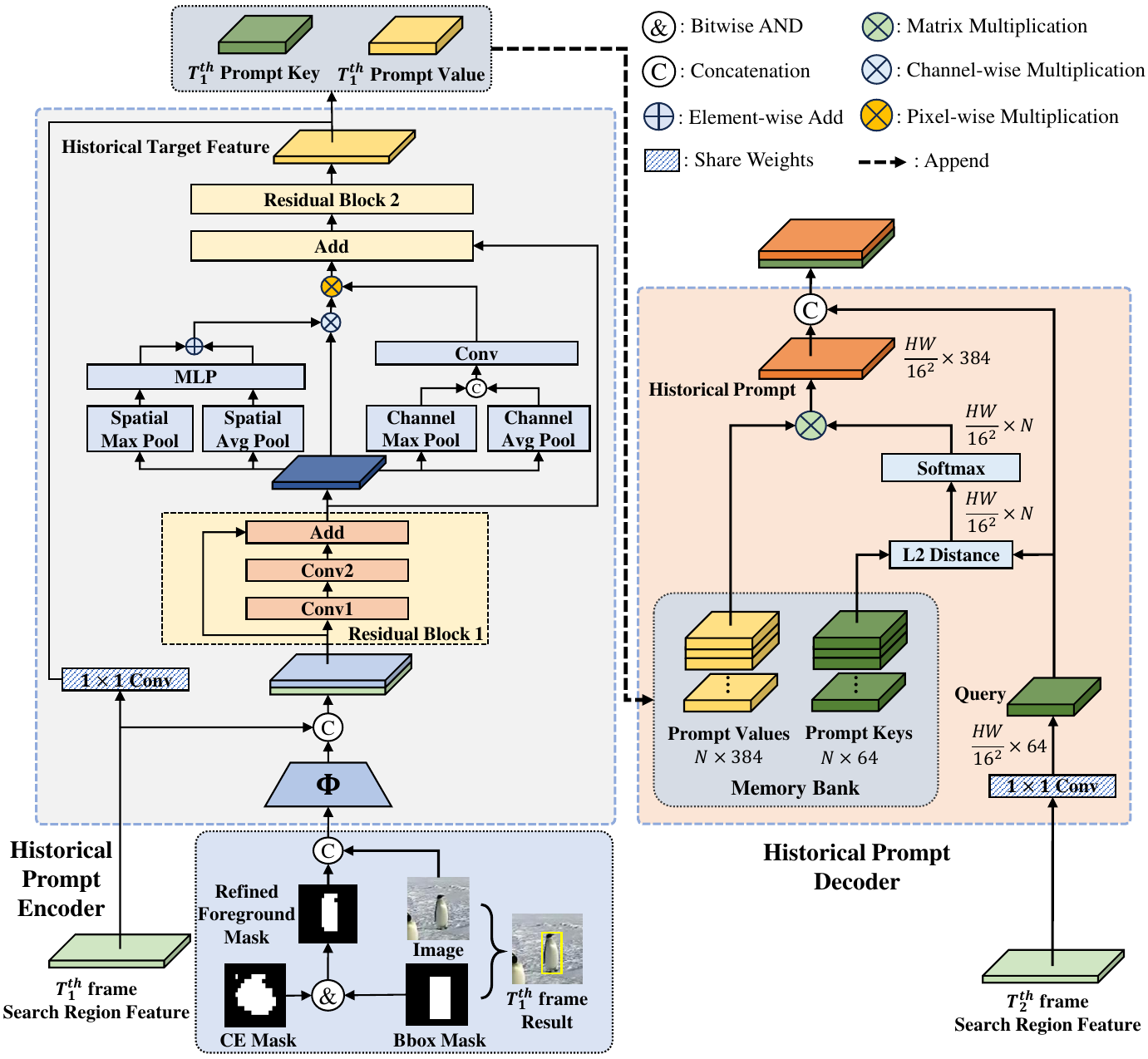}
\caption{
The structure of the historical prompt encoder and the historical prompt decoder. Zoom in for a clearer view. 
%The historical prompt encoder generates historical prompt values, which are stored in memory bank. The historical prompt decoder is used to generate historical prompts suitable for the current search region.
}
\label{fig:prompt}
\end{figure}

As shown in Figure \ref{fig:prompt}, the historical prompt decoder stores the historical prompt values in the form of key-value pairs in the memory bank, utilizes the current search region feature to retrieve historical prompt values, and adaptively aggregates 
 the prompt values to generate customized historical prompt relevant to the current search region.

\textbf{Memory Bank.} We use compressed search region features from the feature extraction network as the prompt key to each prompt value. To reduce computational cost, a $1\times1$ convolutional layer $\mathrm{Conv}_{key}$ is utilized to reduce the channel dimension of $\bm{F}$ from $C_{F}$ to $C_{P_k}$, obtaining the prompt key that is denoted as $\bm{P_{key}} \in \mathbb{R}^{\frac{H}{16}\times \frac{W}{16} \times C_{P_k}}$. After obtaining the prompt values and prompt keys for all the $\frac{H}{16} \times \frac{W}{16}$ positions in the search region of a specific frame, the historical prompt decoder will flatten them along the spatial dimension and incorporate them into the memory bank altogether, which means that the memory bank will add $\frac{HW}{16^2}$ prompt key-value pairs. The memory bank retains at most $T$ tracked frames and is updated every $\tau$ frames, using a first-in-first-out (FIFO) strategy.

\textbf{Decoding.} The decoding process is to adaptively aggregate historical prompt values from the memory bank based on the current search region feature and generate the historical prompt for target prediction. 
%HIQ generates historical information prompts tailored specifically for the present search region. 
Given the current search region feature $\bm{F}$, to ensure that the query aligns precisely with the prompt keys in the memory bank and to reduce computational cost, we also utilize $\mathrm{Conv}_{key}$ to reduce the dimension of $\bm{F}$ to $C_{P_k}$ as the query of current frame, which is denoted as $\bm{Q} \in \mathbb{R}^{\frac{H}{16}\times \frac{W}{16} \times C_{P_k}}$. If the prediction of current frame needs to be encoded as historical target feature and added to the memory bank as historical prompt value, $\bm{Q}$ can be directly employed as the corresponding prompt key without redundant calculations. Assuming there are a total of $\bm{N}$ key-value pairs in the memory bank, the process that the decoder generates historical prompt for current search region can be formulated as follows:

\begin{equation}\label{eq_4}\small
\begin{aligned}
\bm{S_{i,j}} & = -||\bm{P^{\prime}_{key_i}} - \bm{Q_j}||_2^2 \\
 \bm{A_{i,j}} & = \frac{e^{\bm{S_{i,j}}}}{\sum_{n=1}^{N}(e^{\bm{S_{n,j}}})} \\
 \bm{O} & = \bm{A}^T \cdot \bm{P^{\prime}}
\end{aligned}
\end{equation}

\noindent where $\bm{P^{\prime}_{key}} \in \mathbb{R}^{N\times C_{P_k}}$ represents all prompt keys in the memory bank and $\bm{S_{i,j}} \in \mathbb{R}^1$ represents the similarity score between the $i^{th}$ prompt key and the $j^{th}$ query, which is calculated using the negative of L2 distance. $\bm{A} \in \mathbb{R}^{N\times \frac{HW}{16^2}}$ represents the final normalized attention score matrix, $\bm{P^{\prime}} \in \mathbb{R}^{N \times C_P}$ represents 
 all the prompt values in the memory bank, and $\bm{O} \in \mathbb{R}^{\frac{HW}{16^2} \times C_p}$ denotes the historical prompt tailored to the current search region, obtained by weighted aggregation based on matrix $\bm{A}$. After reshaping, $\bm{O}^{\prime} \in \mathbb{R}^{\frac{H}{16} \times \frac{W}{16} \times C_p}$ is concatenated with the compressed search region features and fed into the prediction head. 
 
\section{Experiments}
\subsection{Implemention Details}

\textbf{Model settings.} We employ the base version of Vision Transformer (ViT-B)  \cite{dosovitskiy2020image} in feature extraction network and initialize it using the weights from DropTrack  \cite{Wu_2023_CVPR_dropmae}. Notably, when evaluating on GOT-10k  \cite{huang2019got}, our initialization weights are trained only on GOT-10k.
Throughout the training process, the feature extraction network remains frozen. 
%Thanks to the robust prompts generated by HIP module, our training procedure involves minimal trainable parameters. This allows HIPTrack to effectively sample multiple frames for training within a single training step. 
%When processing template frame and search frames, 
HIPTrack utilizes a template size of $192\times 192$ and a search region size of $384\times 384$. The cropping factors for the template and search region are 2 and 5, respectively. The feature extraction network outputs features with a dimension of $C_F$ that is set to 768. In historical prompt network, we set the values of $C_K$, $C_P$ and $C_{P_k}$ to 256, 384 and 64, respectively.
%Both the query and key features in HIP module are compressed to a dimension of 64. The output feature dimension of the prompt value encoder is set to 384. 
As shown in Table \ref{table:model_settings}, our method requires training only a small number of parameters, while also demonstrating significantly reduced computational complexity during the inference process even when the memory bank is fully utilized. Moreover, our method achieves a speed of 45.3 FPS on one single NVIDIA Tesla V100 GPU, exhibiting a more significant advantage over MACs. This is because during actual tracking, the memory bank is not always full and the encoder is occasionally called at update intervals.

\begin{table}[!h]\small
    \centering
    \caption{ Comparison of our method with other excellent trackers in terms of total parameters, trainable parameters, computational complexity and speed. The speed of all methods is tested on V100.}
    \setlength{\tabcolsep}{0.8mm}
    \begin{tabular}{c|cccc}
    \Xhline{2pt}
        \multirow{2}{*}{\textbf{Method}} & \textbf{Trainable} & \multirow{2}{*}{\textbf{Params(M)}} & \multirow{2}{*}{\textbf{MACs(G)}} & \textbf{Speed}  \\
         & \textbf{Params(M)} & & & \textbf{(FPS)} \\
        \Xhline{1pt}
        \textbf{HIPTrack} & \textbf{34.1} & 120.4 & \textbf{66.9} & \textbf{45.3} \\
        SeqTrack-B$_{384}$\cite{Chen_2023_CVPR_seqtrack} & 88.1 & \textbf{88.1} & 147.9 & 16.8 \\
        ARTrack$_{384}$\cite{Wei_2023_CVPR_autoregressive} & 181.0 & 181.0 & 172.1 & 13.5 \\
        TATrack-L\cite{he2023target_TATrack} & 112.8 & 112.8 & 162.4 & 6.6 \\
    \Xhline{2pt}
    \end{tabular}
    \label{table:model_settings}
\end{table}

\textbf{Datasets.} Following previous works \cite{Cui_2022_MixFormer,ye_2022_joint,Wei_2023_CVPR_autoregressive,Chen_2023_CVPR_seqtrack}, when evaluating the trackers on GOT-10k  \cite{huang2019got}, we only use the \emph{train} splits of GOT-10k for training. Otherwise, we utilize the \emph{train} splits of COCO  \cite{lin2014microsoft}, LaSOT  \cite{fan2019lasot}, GOT-10k  \cite{huang2019got}, and TrackingNet  \cite{muller2018trackingnet}. In each training step, we select six frames from a video. The temporal intervals between adjacent frames are randomly chosen from [1, 70]. Once selected, the six frames are randomly arranged in either chronological or reverse order, with the first frame serving as the template and the remaining frames serving as the search frames. For non-video datasets like COCO, we duplicate the same image six times.

%Assuming that the GT box in each frame is centered at $(x,y)$ and has a width of $w$ and a height of $h$, we crop a square region around $(x,y)$ in the template frame with a cropping factor equal to 2, which means an area of $2 \times (wh)^2$, and use it as the ultimate target template. To prevent the model from learning positional bias  when tracking, we apply jitter to the center point and scale of the ground truth box in the search frame and crop a square region centered around the jittered box  \cite{li2019siamrpn++}. 
%The jittered width $w^{\prime}$ and height $h^{\prime}$ are both multiplied by a factor $a = e^{0.5\tau}$, where $\tau$ follows a uniform distribution between $[0,2]$. The jittered center point coordinates $x^{\prime}$ and $y^{\prime}$ are both added by $b = \frac{(\tau - 0.5) \times \sqrt{w^{\prime}h^{\prime}}}{2}$ . Then we crop a square region centered around$(x^{\prime},y^{\prime})$ with an area of $2\times (w^{\prime}h^{\prime})^2$ to use as the ultimate search region.

\textbf{Loss Function.} In our implementation, we utilize focal loss \cite{lin2017focal} for foreground-background classification and then employ GIoU loss \cite{rezatofighi2019generalized} and L1 loss for bounding box regression. The weighting coefficients for focal loss , GIoU loss, and L1 loss are set as 1.0, 5.0, and 2.0, respectively.

\textbf{Training and Optimization.} Our tracker is implemented using PyTorch 1.10.1. The entire training process is conducted on 4 NVIDIA Tesla V100 GPUs. During training, we set the batch size to 32 and train the model for 100 epochs. In each epoch, we sample 60,000 videos from all the datasets. We use AdamW optimizer with a weight decay of $10^{-4}$ and an initial learning rate of $10^{-4}$. The learning rate is scheduled to decrease to $10^{-5}$ after 80 epochs.

\textbf{Inference.} For the first 10 frames, the historical prompt network utilizes the historical information from the first frame template as memory bank to generate prompts and the historical information of a tracked frame is added to the memory bank every 5 frames. After the initial 10 frames, 
 the historical prompt network utilizes the historical information stored in the memory bank to generate prompts.
 %and the update interval of the memory bank $\tau$ is set to 20. We set the memory bank size $T$ to 150 and preserve the information of the first 10 memory frames during FIFO updates.
 The the update interval of the memory bank $\tau$ is set to 20 and the memory bank size $T$ is set to 150.

\subsection{Comparisions with the State-of-the-Art}

\begin{figure}[!t]
%是可选项 h表示的是here在这里插入，t表示的是在页面的顶部插入
%学术论文里图一般要放在顶部
\centering
\includegraphics[scale=0.48]{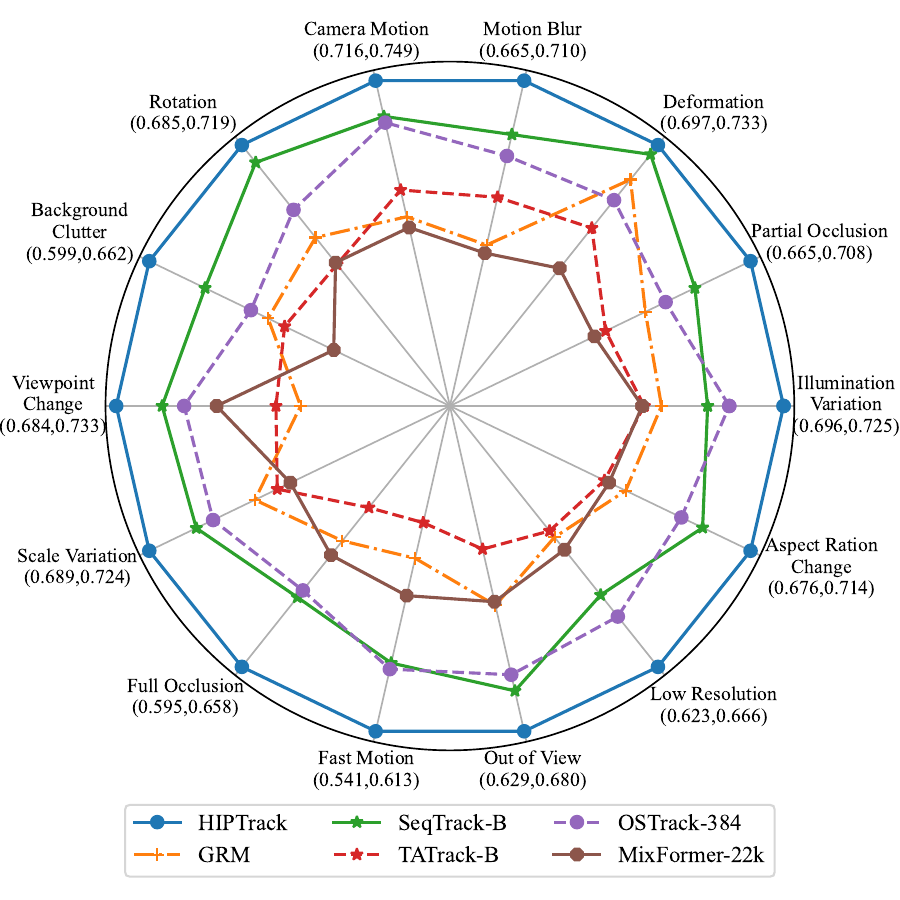}
\caption{The performance of our method compared with other state-of-the-art trackers in terms of AUC across various scenarios in the LaSOT \emph{test} split.}
\label{fig:lasotSubset}
\end{figure}

\textbf{LaSOT}  \cite{fan2019lasot} is a large-scale long-term dataset. Its \emph{test} split consists of 280 sequences, each exceeding 2,500 frames. We evaluate our method on the \emph{test} split of LaSOT, and the results presented in Table  \ref{whole_comparison} show that our approach outperforms current state-of-the-art methods. 
%Notably, even in comparison to trackers such as SeqTrack-B$_{384}$  \cite{Chen_2023_CVPR_seqtrack} and ARTrack$_{384}$  \cite{Wei_2023_CVPR_autoregressive}, which introduce an additional Transformer decoder for autoregressive prediction, our method continues to exhibit superior performance. 
In Figure  \ref{fig:lasotSubset}, we evaluate the performance of our approach across various challenging tracking scenarios. This observation underscores the adaptability and robustness of our approach. 
   
\begin{table*}[!th]\small
    \centering
    \caption{State-of-the-art comparison on LaSOT, GOT-10k and TrackingNet. `*' denotes for trackers trained only with GOT-10k train split. 
 The best two results are highlighted in \textbf{\textcolor{red}{red}} and \textbf{\textcolor{blue}{blue}}, respectively.}
\setlength{\tabcolsep}{1.5mm}
    \label{whole_comparison}
    \begin{tabular}{c|c|cccccccccccc}
        \Xhline{2pt}
        Method & Source & \multicolumn{3}{c}{LaSOT} & \multicolumn{3}{c}{GOT-10k*} & \multicolumn{3}{c}{TrackingNet} \\
        \cmidrule(lr){3-5}\cmidrule(lr){6-8}\cmidrule(lr){9-11} &
         & AUC(\%) & $P_{Norm}$(\%) & $P$(\%) & AO(\%) & $SR_{0.5}$(\%) & $SR_{0.75}$(\%) & AUC(\%) & $P_{Norm}$(\%) & $P$(\%)\\
        \Xcline{1-1}{0.4pt}
        \Xhline{1pt}
        \textbf{HIPTrack} & \textbf{Ours} & \textbf{\textcolor{red}{72.7}} & \textbf{\textcolor{red}{82.9}} & \textbf{\textcolor{red}{79.5}} & \textbf{\textcolor{red}{77.4}} & \textbf{\textcolor{red}{88.0}} & \textbf{\textcolor{red}{74.5}} & \textbf{\textcolor{blue}{84.5}} & \textbf{\textcolor{red}{89.1}} & \textbf{\textcolor{blue}{83.8}} \\
        ROMTrack-384 \cite{Cai_2023_ICCV_ROMTrack} & ICCV23 & 71.4 & 81.4 & 78.2 & 74.2 & 84.3 & 72.4 & 84.1 & \textbf{\textcolor{blue}{89.0}} & 83.7 \\
        DropTrack  \cite{Wu_2023_CVPR_dropmae} & CVPR23 & 71.8 & \textbf{\textcolor{blue}{81.8}} & 78.1 & \textbf{\textcolor{blue}{75.9}} & \textbf{\textcolor{blue}{86.8}} & 72.0 & 84.1 & 88.9 & - \\
        ARTrack$_{384}$  \cite{Wei_2023_CVPR_autoregressive} & CVPR23 & \textbf{\textcolor{blue}{72.6}} & 81.7 & \textbf{\textcolor{blue}{79.1}} & 75.5 & 84.3 & \textbf{\textcolor{blue}{74.3}} & \textbf{\textcolor{red}{85.1}} & \textbf{\textcolor{red}{89.1}} & \textbf{\textcolor{red}{84.8}} \\
        SeqTrack-B$_{384}$   \cite{Chen_2023_CVPR_seqtrack} & CVPR23 & 71.5 & 81.1 & 77.8 & 74.5 & 84.3 & 71.4 & 83.9 & 88.8 & 83.6 \\
        GRM   \cite{Gao_2023_CVPR_GRM} & CVPR23 & 69.9 & 79.3 & 75.8 & 73.4 & 82.9 & 70.4 & 84.0 & 88.7 & 83.3 \\
        TATrack-B  \cite{he2023target_TATrack} & AAAI23 & 69.4 & 78.2 & 74.1 & 73.0 & 83.3 & 68.5 & 83.5 & 88.3 & 81.8 \\
        CTTrack  \cite{Song_2023_AAAI_CTTrack} & AAAI23 & 67.8 & 77.8 & 74.0 & 71.3 & 80.7 & 70.3 & 82.5 & 87.1 & 80.3 \\
        OSTrack$_{384}$  \cite{ye_2022_joint} & ECCV22 & 71.1 & 81.1 & 77.6 & 73.7 & 83.2 & 70.8 & 83.9 & 88.5 & 83.2 \\
        SimTrack  \cite{chen2022backbone} & ECCV22 & 70.5 & 79.7 & - & 69.8 & 78.8 & 66.0 & 83.4 & 87.4 & - \\
        MixFormer-22K  \cite{Cui_2022_MixFormer} & CVPR22 & 69.2 & 78.7 & 74.7 & 70.7 & 80.0 & 67.8 & 83.1 & 88.1 & 81.6 \\
        SBT  \cite{Xie_2022_Correlation} & CVPR22 & 66.7 & - & 71.7 & 70.4 & 80.8 & 64.7 & - & - & - \\
        AiATrack  \cite{gao2022aiatrack} & ECCV22 & 69.0 & 79.4 & 73.8 & 69.6 & 80.0 & 63.2 & 82.7 & 87.8 & 80.4 \\
        SwinTrack  \cite{lin2022swintrack} & NIPS22 & 71.3 & - & 76.5 & 72.4 & - & 67.8 & 84.0 & - & 82.8 \\
        SparseTT  \cite{fu2022sparsett} & IJCAI22 & 66.0 & 74.8 & 70.1 & 69.3 & 79.1 & 63.8 & 81.7 & 86.6 & 79.5 \\
        STARK  \cite{yan2021learning} & ICCV21 &  67.1 & 77.0 & - & 68.8 & 78.1 & 64.1 & 82.0 & 86.9 & -\\
        
        \Xhline{2pt}
    \end{tabular} 
\end{table*}

\textbf{LaSOT$_{\mathrm{ext}}$}  \cite{fan2021lasot} comprises 1,500 video sequences and 15 distinct target categories that have no overlaps with those in the LaSOT  \cite{fan2019lasot} dataset. Table  \ref{comparison_lasto_extf} demonstrates a significant superiority of our approach over SeqTrack-B$_{384}$ 
%(\textbf{+2.5\%AUC})
, ARTrack$_{384}$ 
%(\textbf{+1.1\%AUC})
, and OSTrack$_{384}$
%(\textbf{+2.5\%AUC}) 
on LaSOT$_{\mathrm{ext}}$.

\begin{table}[]\small
    \centering
    \caption{The performance of our method and other state-of-the-art trackers on LaSOT$_{\mathrm{ext}}$. The best two results are highlighted in \textbf{\textcolor{red}{red}} and \textbf{\textcolor{blue}{blue}}.}
    \setlength{\tabcolsep}{2mm}
    \begin{tabular}{c|ccc}
    \Xhline{2pt}
        \textbf{Method} & AUC(\%) & $P_{Norm}$(\%) & $P$(\%)  \\
        \Xhline{1pt}
        \textbf{HIPTrack} & \textbf{\textcolor{red}{53.0}} & \textbf{\textcolor{red}{64.3}} & \textbf{\textcolor{red}{60.6}} \\
        SeqTrack-B$_{384}$  \cite{Chen_2023_CVPR_seqtrack} & 50.5 & 61.6 & 57.5 \\
        ARTrack$_{384}$  \cite{Wei_2023_CVPR_autoregressive} & \textbf{\textcolor{blue}{51.9}} & \textbf{\textcolor{blue}{62.0}} & \textbf{\textcolor{blue}{58.5}} \\
        OSTrack$_{384}$  \cite{ye_2022_joint} & 50.5 & 61.3 & 57.6\\ 
        AiATrack  \cite{gao2022aiatrack} & 47.7 & 55.6 & 55.4 \\
        SwinTrack  \cite{lin2022swintrack} & 49.1 & - & 55.6 \\
        ToMP  \cite{mayer2022transforming} & 45.9 & - & - \\
        KeepTrack  \cite{mayer2021learning} & 48.2 & - & - \\
        LTMU  \cite{dai2020high} & 41.4 & 49.9 & 47.3 \\
        DiMP  \cite{bhat2019learning} & 39.2 & 47.6 & 45.1 \\
    \Xhline{2pt}
    \end{tabular}
    \label{comparison_lasto_extf}
\end{table}

\textbf{GOT-10k}  \cite{huang2019got} contains 9,335 sequences for training and 180
sequences for testing. GOT-10k only allows trackers to be trained using the \emph{train} split.
We follow this protocol to train our method on the \emph{train} split and test it on the
\emph{test} split. Table  \ref{whole_comparison} indicates that 
our method surpasses all current state-of-the-art methods as well,
%non-autoregressive methods as well as the autoregressive method SeqTrack  \cite{Chen_2023_CVPR_seqtrack}, 
showcasing the robust capability of our approach in extracting historical information and generating prompts even for unknown categories.

\textbf{TrackingNet}  \cite{muller2018trackingnet} is a large-scale dataset whose \emph{test} split
includes 511 sequences covering various object classes and
tracking scenes. We report the performance of our method on the \emph{test} split of TrackingNet. Table  \ref{whole_comparison} demonstrates that our method outperforms all current non-autoregressive methods as well as the autoregressive method SeqTrack  \cite{Chen_2023_CVPR_seqtrack}.

\textbf{UAV123}  \cite{mueller2016benchmark} is a low altitude aerial dataset taken by drones, including 123 sequences, with an average of 915 frames per sequence. The results in Table  \ref{table:uav,nfs,otb} indicate that our approach rivals to the current state-of-the-art methods on UAV123. The reason for not achieving more significant advancements could be attributed to the relatively smaller target scales in UAV123, which may lead to a decrease in accuracy when using masks to describe the target position.

\textbf{NfS}  \cite{Galoogahi_2017_ICCV_nfs} comprises 100 video sequences, totaling 380,000 video frames.  We experiment on the 30FPS version of NfS. The results in Table  \ref{table:uav,nfs,otb} demonstrate that our approach outperforms all current state-of-the-art approaches.

\textbf{OTB2015}  \cite{otb2015} is a classical testing dataset in visual tracking. It contains 100 short-term tracking sequences covering 11 common challenges, such as target deformation, occlusion and scale variation. The results in Table  \ref{table:uav,nfs,otb} demonstrate that our approach surpasses current state-of-the-art methods on 
 OTB2015 as well.

\begin{table}[h]\small
    \centering
    \caption{The performance of our method and other state-of-the-art trackers on UAV123, NfS and OTB2015 in terms of AUC metrics. The best two results are highlighted in \textbf{\textcolor{red}{red}} and \textbf{\textcolor{blue}{blue}}.}
    \setlength{\tabcolsep}{2mm}
    \begin{tabular}{c|ccc}
    \Xhline{2pt}
        \textbf{Method} & \textbf{UAV123} & \textbf{NfS} & \textbf{OTB2015}  \\
        \Xhline{1pt}
        \textbf{HIPTrack} & \textbf{\textcolor{blue}{70.5}} & \textbf{\textcolor{red}{68.1}} & \textbf{\textcolor{red}{71.0}} \\
        ARTrack$_{384}$  \cite{Wei_2023_CVPR_autoregressive} & \textbf{\textcolor{blue}{70.5}} & 66.8 & - \\
        AiATrack  \cite{gao2022aiatrack} & \textbf{\textcolor{red}{70.6}} & \textbf{\textcolor{blue}{67.9}} & 69.6 \\
        CTTrack-B  \cite{Song_2023_AAAI_CTTrack} & 68.8 & - & - \\
        SeqTrack-B$_{384}$  \cite{Chen_2023_CVPR_seqtrack} & 68.6 & 66.7 & - \\
        DropTrack  \cite{Wu_2023_CVPR_dropmae} & - & - & 69.6 \\
        MixFormer-L  \cite{Cui_2022_MixFormer} & 69.5 & - & - \\
        KeepTrack  \cite{mayer2021learning} & 69.7 & 66.4 & \textbf{\textcolor{blue}{70.9}} \\
        STARK  \cite{yan2021learning} & 69.1 & - & 68.5 \\
        
    \Xhline{2pt}
    \end{tabular}
    \label{table:uav,nfs,otb}
\end{table}

\subsection{Ablation Studies}

\textbf{Generalization ability of Historical Prompt Network.} 
%HIP module is an effective tool that provides historical information prompts for existing trackers that use template-search region similarity matching for tracking. It is important to note that the training process does not require retraining the parameters of the backbone. 
To test the general applicability of the historical prompt network, we integrate it into Transformer-based one-stream trackers DropTrack \cite{Wu_2023_CVPR_dropmae} and OSTrack  \cite{xu2019joint}, as well as SiamFC++  \cite{xu2020siamfc++} that employs an explicit Siamese  structure. Due to the limitations of convolutional network-based SiamFC++ in performing candidate elimination, 
 the historical prompt network in SiamFC++ is exclusively constructed using historical predicted bounding boxes to form masks without CE masks. The experimental results, as depicted in Table  \ref{table_Generalization}, 
demonstrate that our proposed historical prompt network significantly improves the performance of  existing methods that follow Siamese paradigm. The historical prompt network can still have a great impact on SiamFC++ without introducing precise position information, possibly due to the limited ability of convolutional network-based trackers to comprehensively represent targets. Through the use of richer historical information, the historical prompt network can help refine the representation of the target.

%demonstrate that our proposed Historical Prompt Network consistently enhances the performance of various existing methods. Additionally, we also observe that Historical Prompt Network has a more significant performance enhancement effect on convolutional network-based trackers. This could be attributed to the inherent limitations of convolutional network-based trackers in representing targets comprehensively. The introduction of richer historical information through Historical Prompt Network is likely to contribute to refining the representation of the target.

\begin{table}[!h]\small
    \centering
    \caption{A performance comparison of existing trackers and their integration with our proposed historical prompt network on the GOT-10k \emph{test} set.}
    \setlength{\tabcolsep}{2.2mm}
    \begin{tabular}{c|ccc}
    \Xhline{2pt}
        \textbf{Method} & \textbf{AO(\%)} & $\mathbf{SR_{0.5}}(\%)$ & $\mathbf{SR_{0.75}}(\%)$  \\
        \Xhline{1pt}
        DropTrack  & 75.9 & 86.8 & 72.0 \\
        DropTrack w/ HIP & \textbf{77.4} & \textbf{88.0} & \textbf{74.5} \\
    \Xhline{1pt}
        OSTrack & 73.7 & 83.2 & 70.8 \\
        OSTrack w/ HIP & \textbf{75.4} & \textbf{85.0} & \textbf{73.7} \\
    \Xhline{1pt}
        SiamFC++ & 59.5 & 69.5 & 47.9 \\
        SiamFC++ w/ HIP & \textbf{61.0} & \textbf{71.5} & \textbf{49.6} \\
    \Xhline{2pt}
    \end{tabular}
    \label{table_Generalization}
\end{table}

\textbf{The Number of sampled search frames.} Training HIPTrack involves sampling multiple frames as search frames from a video. However, due to resource limitations, it is not feasible to sample an excessive number of search frames at once. Hence, we conducted an analysis on the impact of the number of sampled search frames on the final tracking performance. As depicted in Table \ref{table3_searchFrames}, the tracker achieves optimal performance with a sample size of 5 frames. A larger number of sampled frames may require a larger model size and training epochs to match, and the scaling capability of the historical prompt network remains to be tested.
%The overall trend indicates that the more search frames are sampled, HIPTrack exhibits stronger capabilities in extracting and exploring target historical features, which also suggests that our method still has room for performance improvement.

\begin{table}[!h]\small
    \centering
    \caption{The performance of our method on the \emph{test} split of
LaSOT when setting different number of sampled search frames.}
    \setlength{\tabcolsep}{3mm}
    \begin{tabular}{c|ccccc}
    \Xhline{2pt}
        \textbf{Number} & \textbf{2} & \textbf{3} & \textbf{4} & \textbf{5} & \textbf{6} \\
        \Xhline{1pt}
        \textbf{AUC}(\%) & 72.1 & 72.5 & 72.4 & \textbf{72.7} & 72.4 \\
        $\mathbf{P_{Norm}}$(\%) & 82.3 & 82.7 & 82.5 & \textbf{82.9} & 82.6 \\
        $\mathbf{P}$(\%) & 78.9 & 79.2 & 79.1 & \textbf{79.5} & 79.2 \\
        
    \Xhline{2pt}
    \end{tabular}
    \label{table3_searchFrames}
\end{table}

\textbf{Ablation Studies on Historical Prompt Network.} Historical Prompt Network simultaneously integrates historical positional information and historical visual feature of the target. In order to assess the individual impacts of these two categories of information on HIPTrack, we conducted separate ablation experiments by removing historical refined foreground masks and historical search region features as inputs to the historical prompt network. The results in the first, second, and sixth rows of Table  \ref{table_other_ablation} suggest that incorporating both refined foreground masks and search region features are beneficial for tracking.  

We also examine the importance of CE mask that describes more precise historical positional information. The results in the third and sixth rows of Table  \ref{table_other_ablation} reveal that using candidate elimination to filter background patches to create a refined mask for introducing accurate positional information is essential. Moreover, we explore the necessity of including background search region image patches in the memory bank. The fourth and sixth rows of Table  \ref{table_other_ablation} 
indicate that including all feature maps yields better results, likely because the constructed mask may not completely cover the target region. In fifth row, we replace the L2 distance of attention calculation with dot product, and the comparison with sixth row shows that L2 distance is better to calculate attention score because our query and key reside in exactly the same feature space. 
%Additional ablation studies regarding the design of the historical prompt network can be found in the supplementary materials.

\begin{figure}[!t]
%是可选项 h表示的是here在这里插入，t表示的是在页面的顶部插入
%学术论文里图一般要放在顶部
\centering
\includegraphics[scale=0.38]{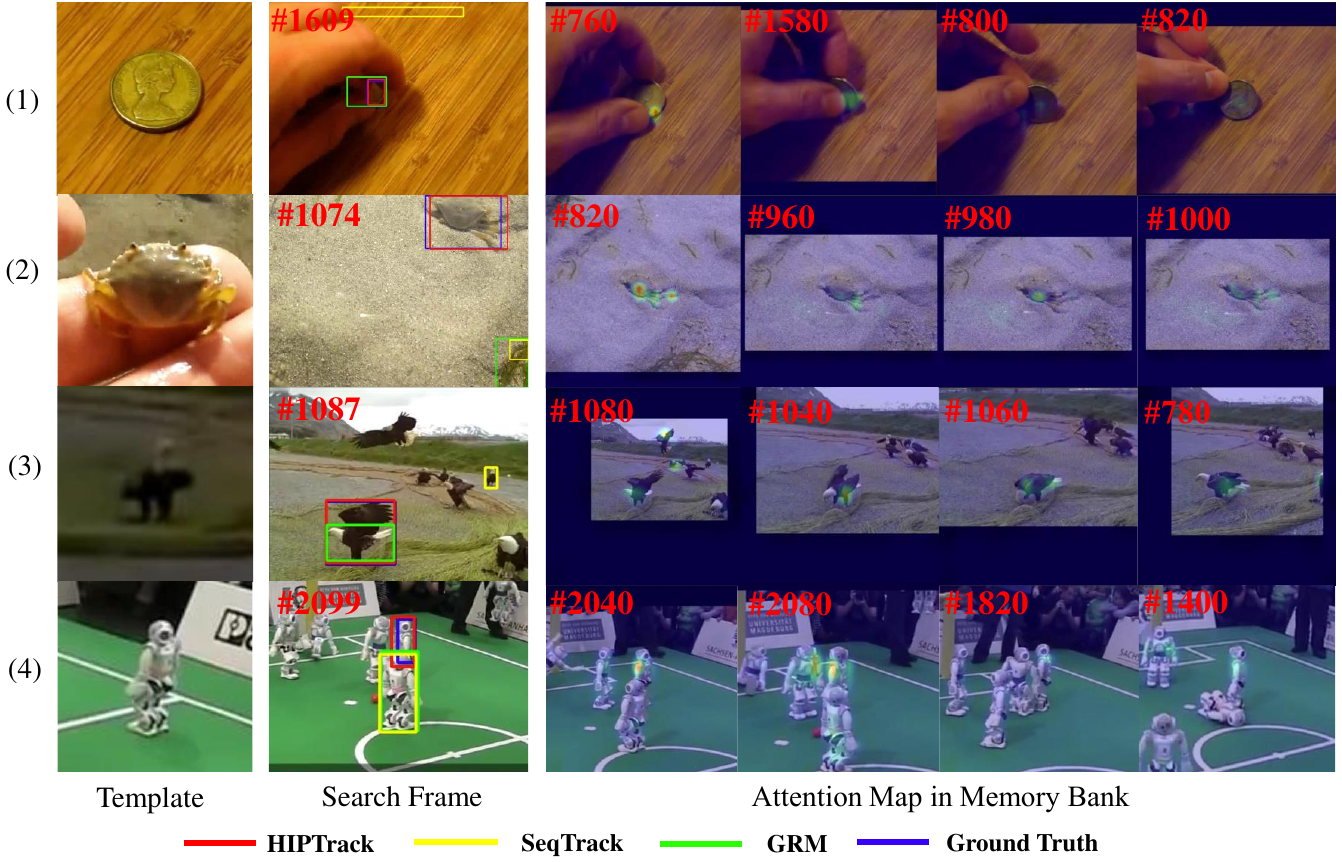}
\caption{Visualization results of memory bank attention maps after prolonged tracking. Zoom in for a clearer view.}
\label{fig:qualitative}
\end{figure}

\begin{table}[]\small
    \centering
    \caption{Ablation studies on refined foreground masks, search region features, candidate elimination, incorporating background region features into the memory bank and L2 attention.}
    %caption 太长了，精简一下，有哪几个模块就说： Ablation study on ....
    \setlength{\tabcolsep}{1mm}
    \begin{tabular}{c|ccccc|c}
    \Xhline{2pt}
        \textbf{\#} & \textbf{Mask} & \textbf{Feature} & \textbf{CE} & \textbf{Background} & \textbf{L2} &\textbf{LaSOT AUC(\%)} \\
        \Xhline{1pt}
        \textbf{1} & \ding{52} & \ding{56} & \ding{52} & \ding{52} & \ding{52} & 71.2 \\
        \textbf{2} &  \ding{56} & \ding{52} & \ding{56} & \ding{52} & \ding{52} & 71.5 \\
        \textbf{3} & \ding{52} & \ding{52} & \ding{56} & \ding{52} & \ding{52} & 72.1 \\
        \textbf{4} & \ding{52} & \ding{52} & \ding{52} &  \ding{56} & \ding{52} & 72.3 \\
        \textbf{5} & \ding{52} & \ding{52} & \ding{52} &  \ding{52} & \ding{56} & 72.3 \\
        \textbf{6} & \ding{52} & \ding{52} & \ding{52} &  \ding{52} & \ding{52} & \textbf{72.7} \\
    \Xhline{2pt}
    \end{tabular}
    \label{table_other_ablation}
\end{table}

\subsection{Qualitative Study}
In Figure \ref{fig:qualitative}, we visualize the comparative tracking results of different methods after prolonged tracking, considering changes in the target due to occlusion, deformation and scale variation. We also visualize the attention maps in memory bank.
The first column represents the template of the first frame in video, the second column shows the comparison of tracking results from different methods, and the third column presents the visualization of attention maps. Figure \ref{fig:qualitative} demonstrates that our method can effectively query and aggregate historical feature of the target during the tracking process. Compared with other methods like GRM \cite{Gao_2023_CVPR_GRM} that relies solely on the first frame template, and the autoregressive method SeqTrack \cite{Chen_2023_CVPR_seqtrack} that does not fully utilize historical visual features of the target, our approach achieves more accurate tracking results.

%\textbf{Visualization of Attention Maps on Historical Information.} We visualized the attention maps of the memory pool in the tracking process of HIPTrack. Specifically, we visualized the three historical search frames that exhibited the highest attention scores collectively within the search region.
\section{Conclusion}
In this work, we identify that trackers that follow Siamese paradigm exhibit significant performance improvements when provided with precise and updated historical information.
%the primary bottleneck limiting the performance improvement of current Transformer-based one-stream trackers lies in their inability to introduce historical information. 
Based on this observation, we propose the historical prompt network that effectively leverages the visual features and positional information of tracked frames to encode the historical target feature, while adaptively generates historical prompts for subsequent frames to enhance tracking accuracy. Our proposed HIPTrack, which features the historical prompt network as its core module, achieves state-of-the-art performance with a small number of parameters that need to be  trained. The historical prompt network can also serve as a
 plug-and-play component to improve the performance of
 current trackers.

\noindent \textbf{Acknowledgements.} This paper was supported by National Natural Science Foundation of China under grants
U20B2069 and 62176017.

%of the target introduces historical positional information through the construction of refined target masks, and simultaneously incorporates historical visual information to generate dedicated historical information prompts for each search region. 
%The entire process introduces only a small number of parameters and requires no training of the backbone, making it readily applicable to existing trackers. Building upon HIP module, we introduce HIPTrack, achieving state-of-the-art performance on datasets such as LaSOT and GOT-10k.
%\input{sec/2_formatting}
%\input{sec/3_finalcopy}

% WARNING: do not forget to delete the supplementary pages from your submission 
\clearpage
\maketitlesupplementary

In this supplementary material, we present additional ablation studies on the historical prompt encoder and the historical prompt decoder
in Section \ref{sec1} to demonstrate the effectiveness of our design. In Section \ref{sec2}, we provide more comprehensive performance comparisons between our proposed HIPTrack and other trackers on LaSOT \cite{fan2019lasot} \emph{test} split, as well as their performance in different complex scenarios within LaSOT. In Section \ref{sec3}, we provide more qualitative visualization analyses.

\section{Further Analyses}\label{sec1}

\begin{figure*}[!t]
\centering
{
\begin{minipage}{4.1cm}
\centering
\includegraphics[scale=0.23]{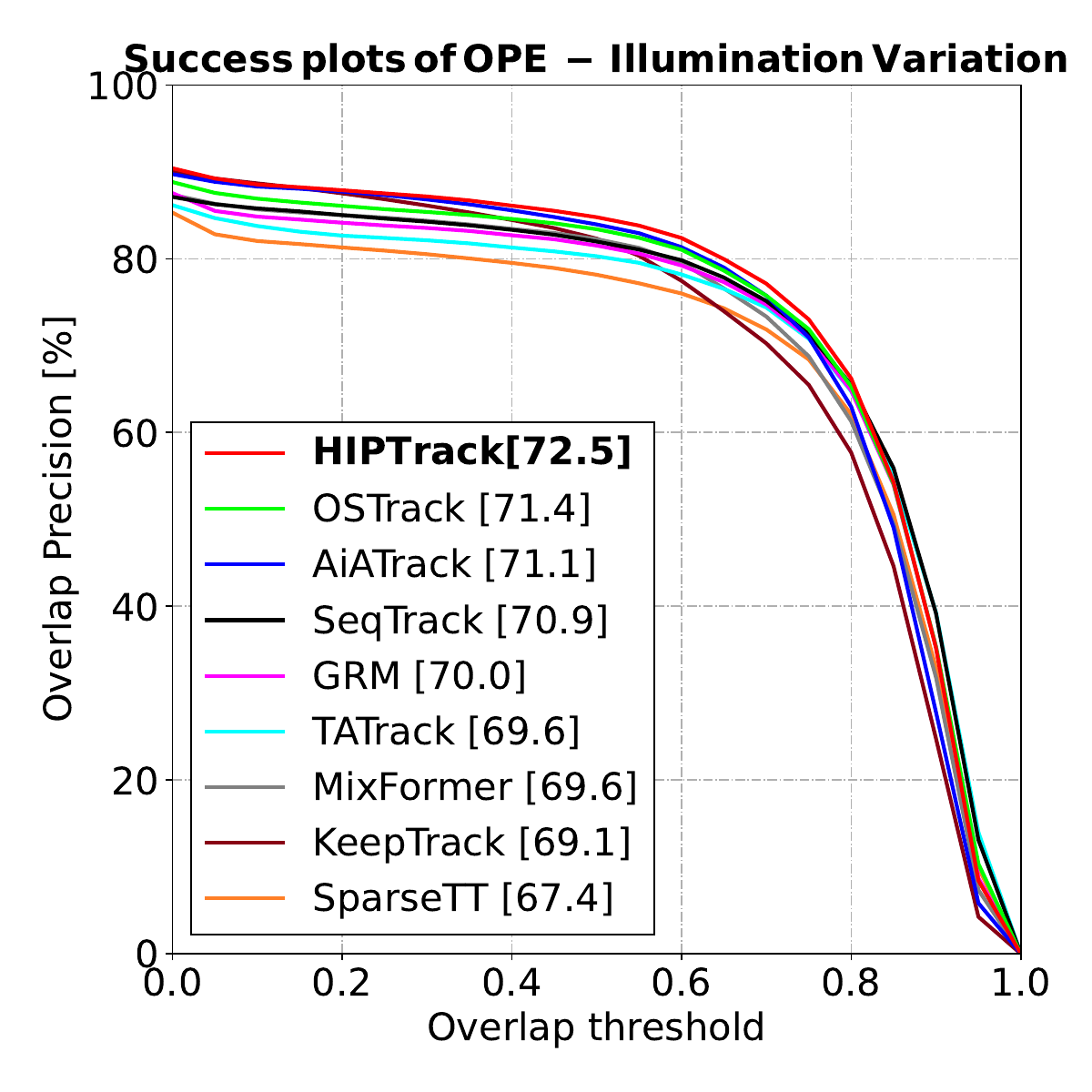} %以这幅图的0.5倍大小输出
\end{minipage}
}
{
\begin{minipage}{4.1cm}
\centering
\includegraphics[scale=0.23]{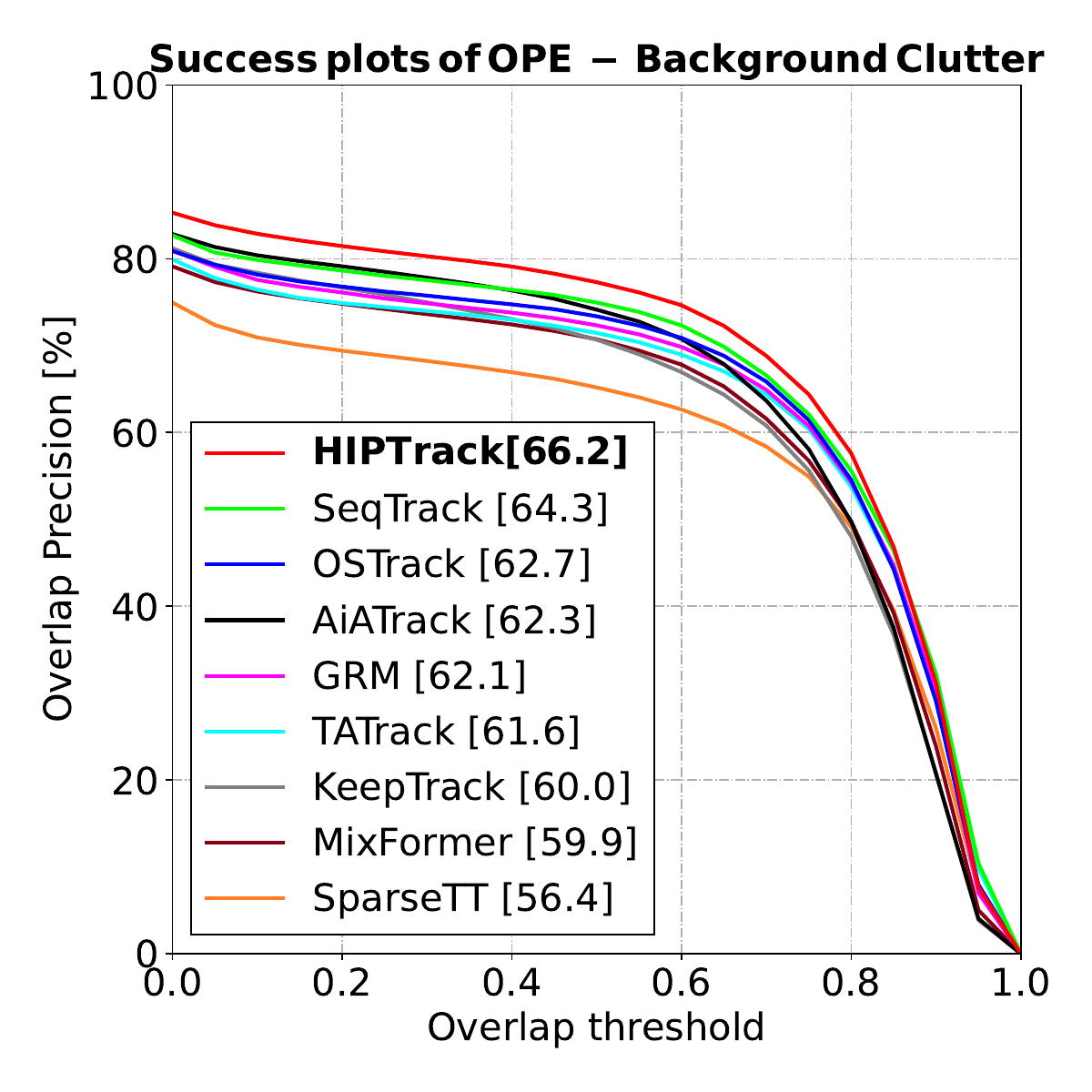}%以这幅图的0.5倍大小输出
\end{minipage}
}
{
\begin{minipage}{4.1cm}
\centering
\includegraphics[scale=0.23]{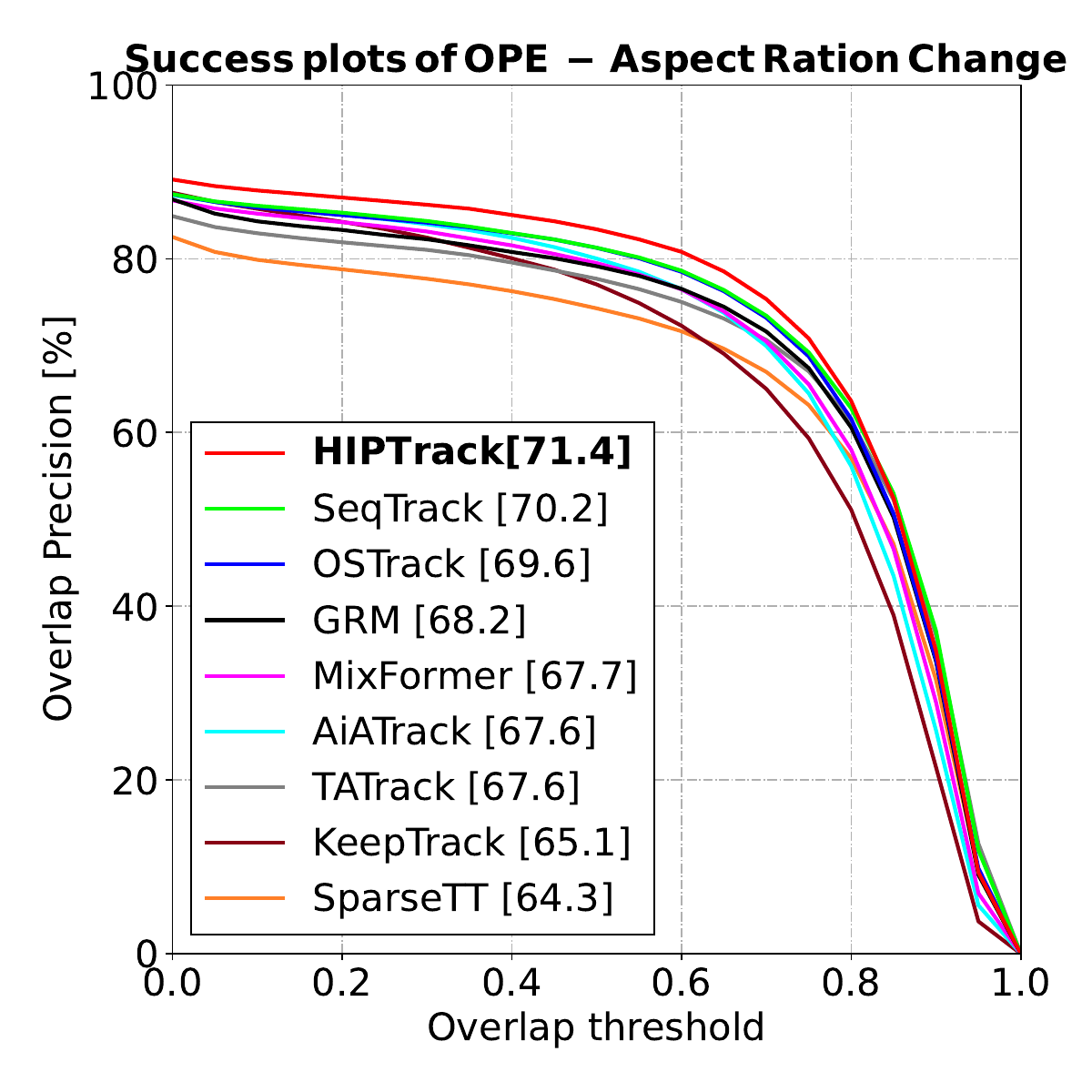}%以这幅图的0.5倍大小输出
\end{minipage}
}
{
\begin{minipage}{4.1cm}
\centering
\includegraphics[scale=0.23]{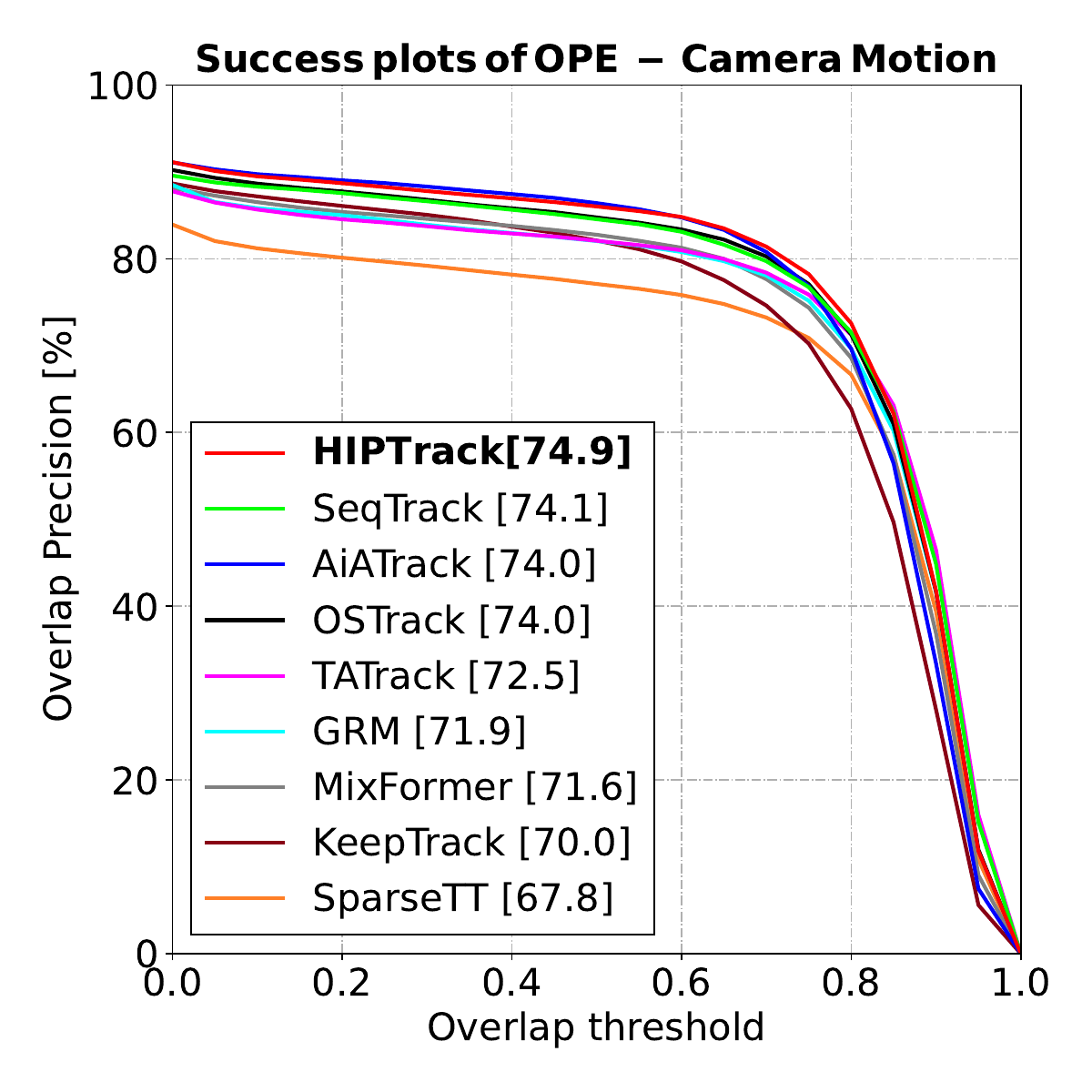}%以这幅图的0.5倍大小输出
\end{minipage}
}

{
\begin{minipage}{4.1cm}
\centering
\includegraphics[scale=0.23]{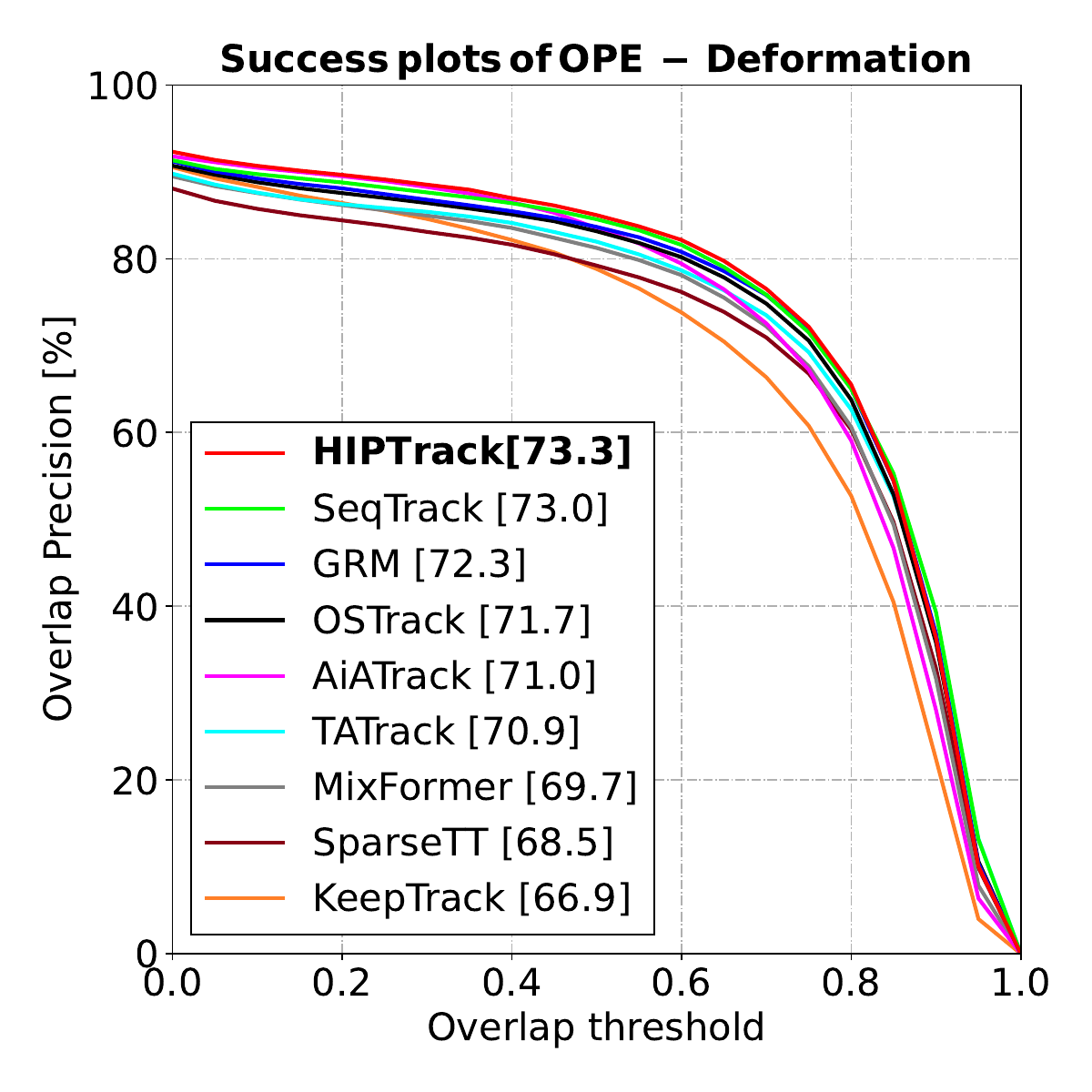}%以这幅图的0.5倍大小输出
\end{minipage}
}
{
\begin{minipage}{4.1cm}
\centering
\includegraphics[scale=0.23]{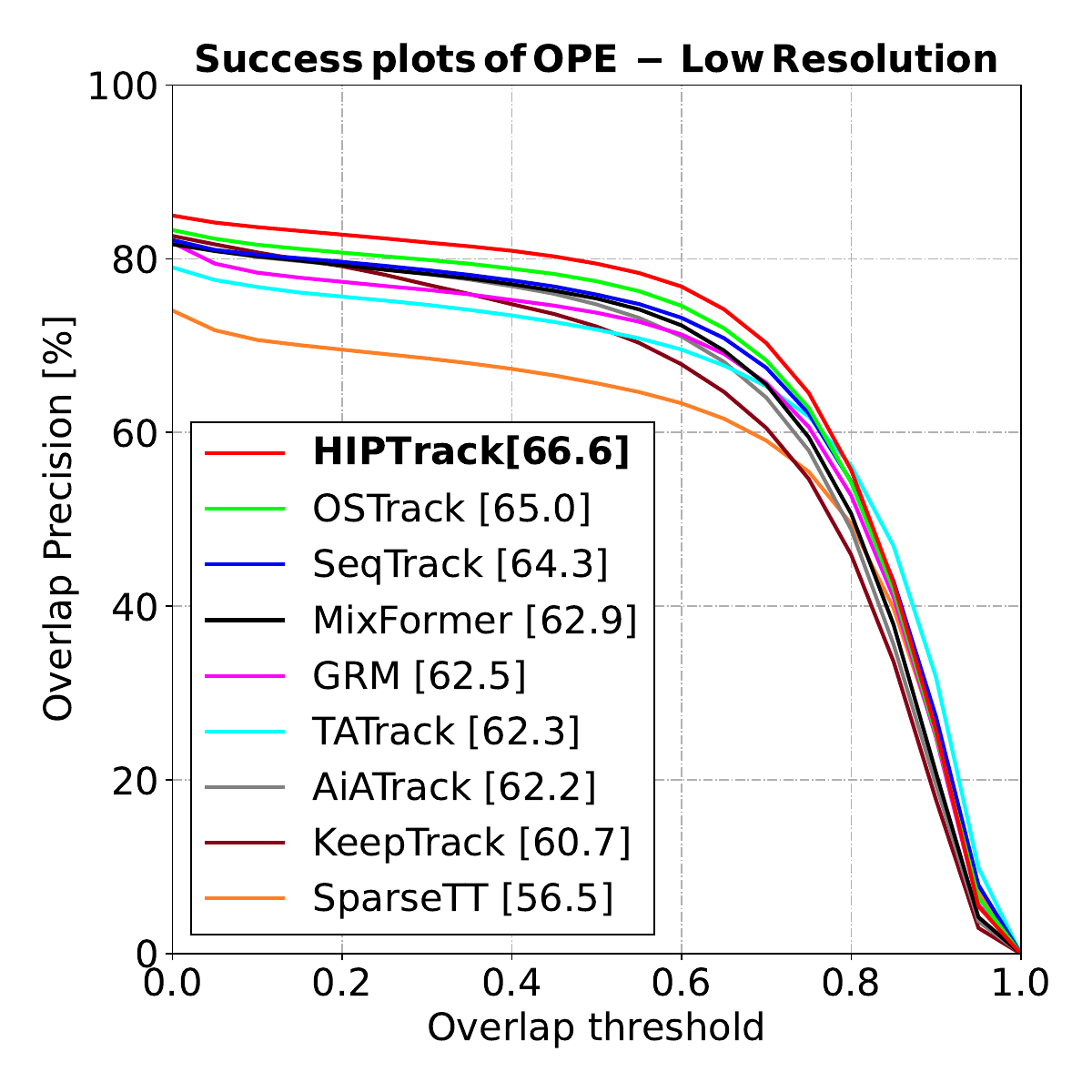}%以这幅图的0.5倍大小输出
\end{minipage}
}
{
\begin{minipage}{4.1cm}
\centering
\includegraphics[scale=0.23]{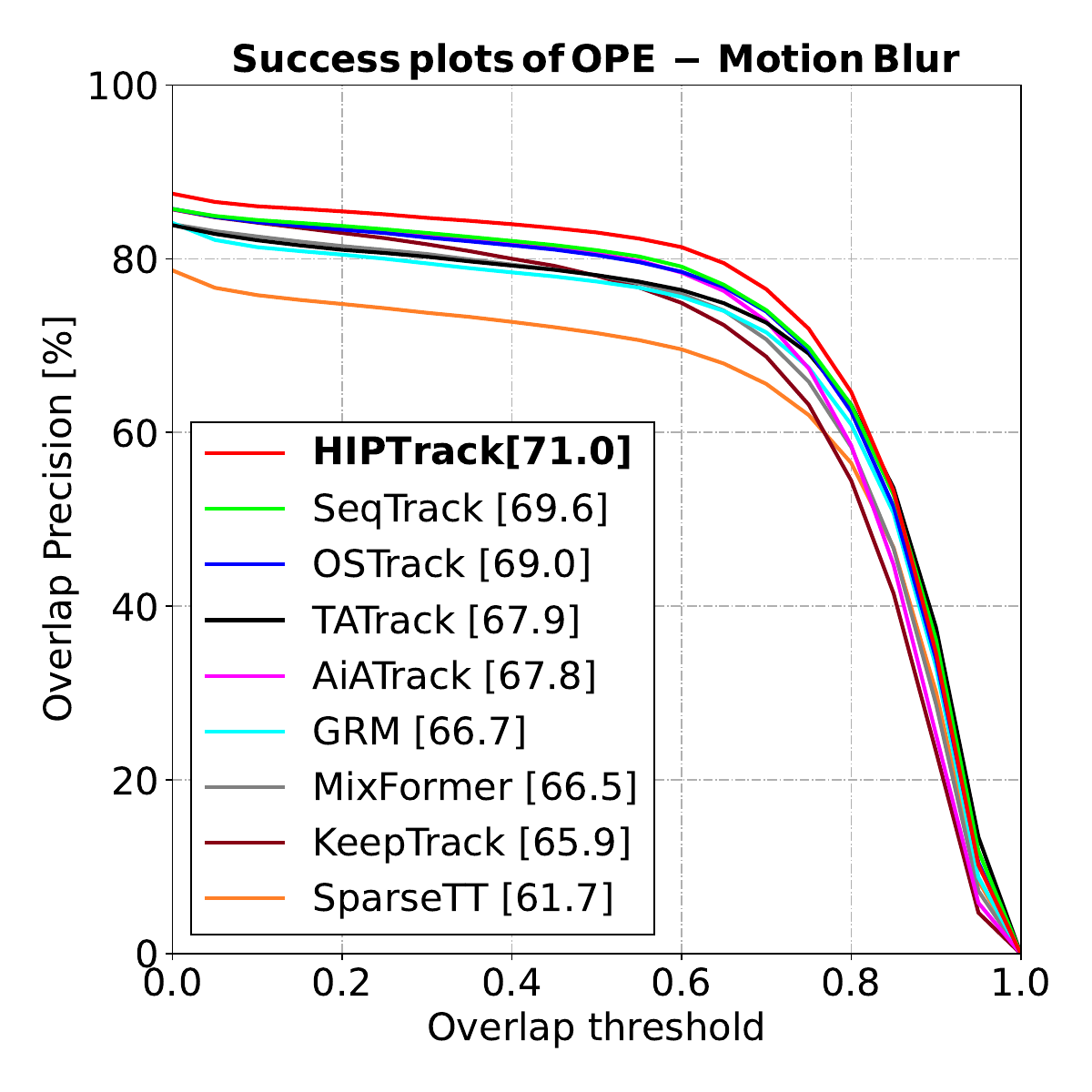}%以这幅图的0.5倍大小输出
\end{minipage}
}
{
\begin{minipage}{4.1cm}
\centering
\includegraphics[scale=0.23]{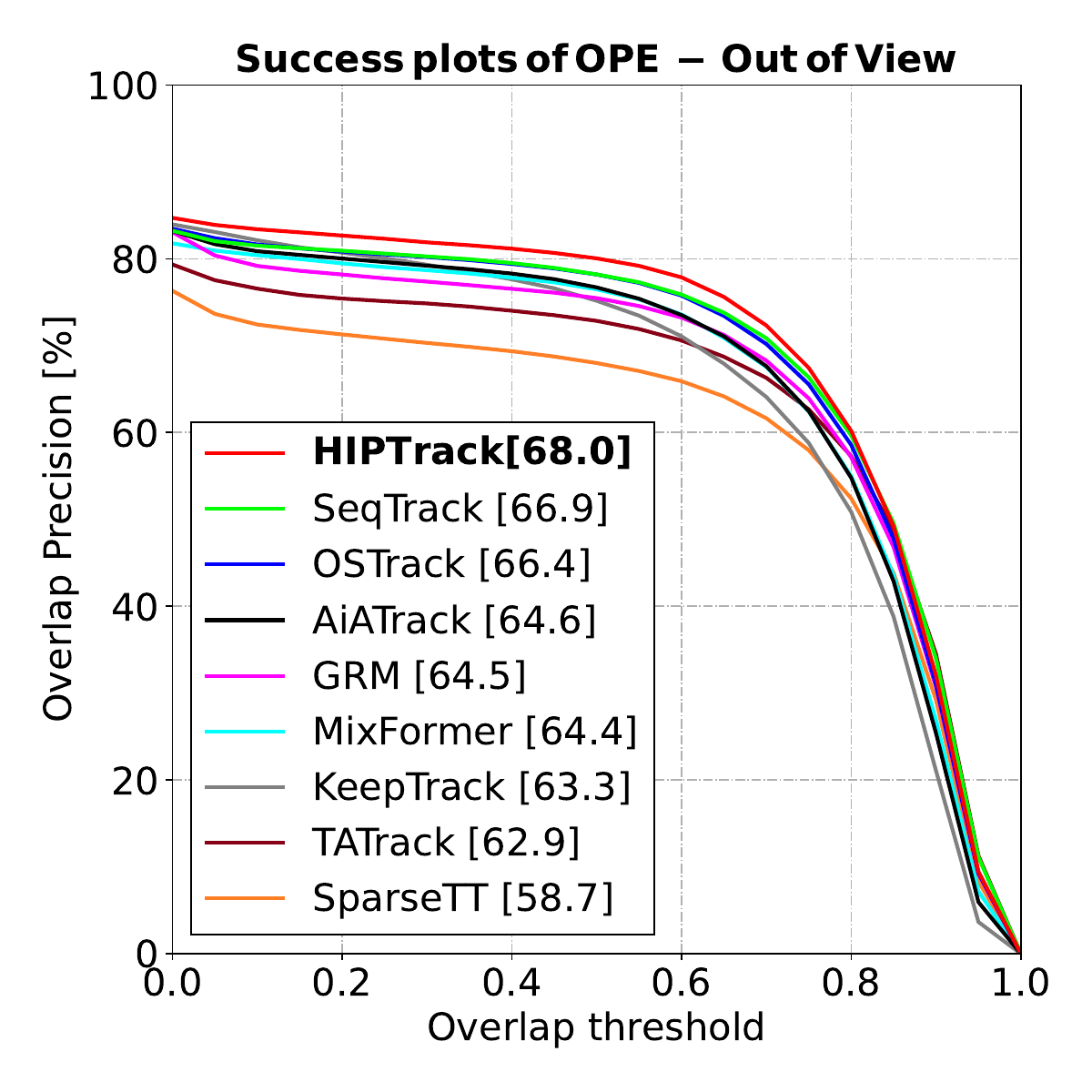}%以这幅图的0.5倍大小输出
\end{minipage}
}

{
\begin{minipage}{4.1cm}
\centering
\includegraphics[scale=0.23]{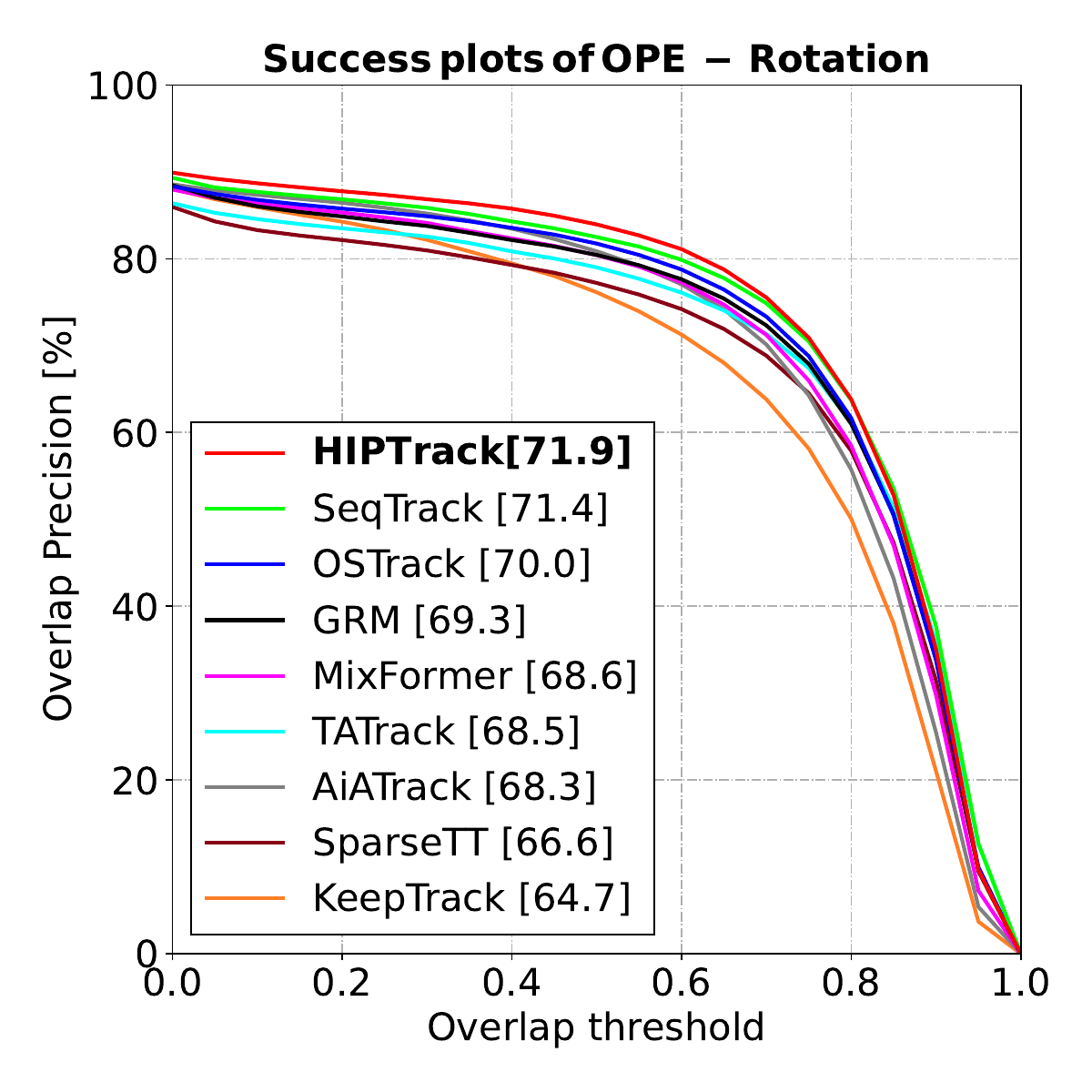}%以这幅图的0.5倍大小输出
\end{minipage}
}
{
\begin{minipage}{4.1cm}
\centering
\includegraphics[scale=0.23]{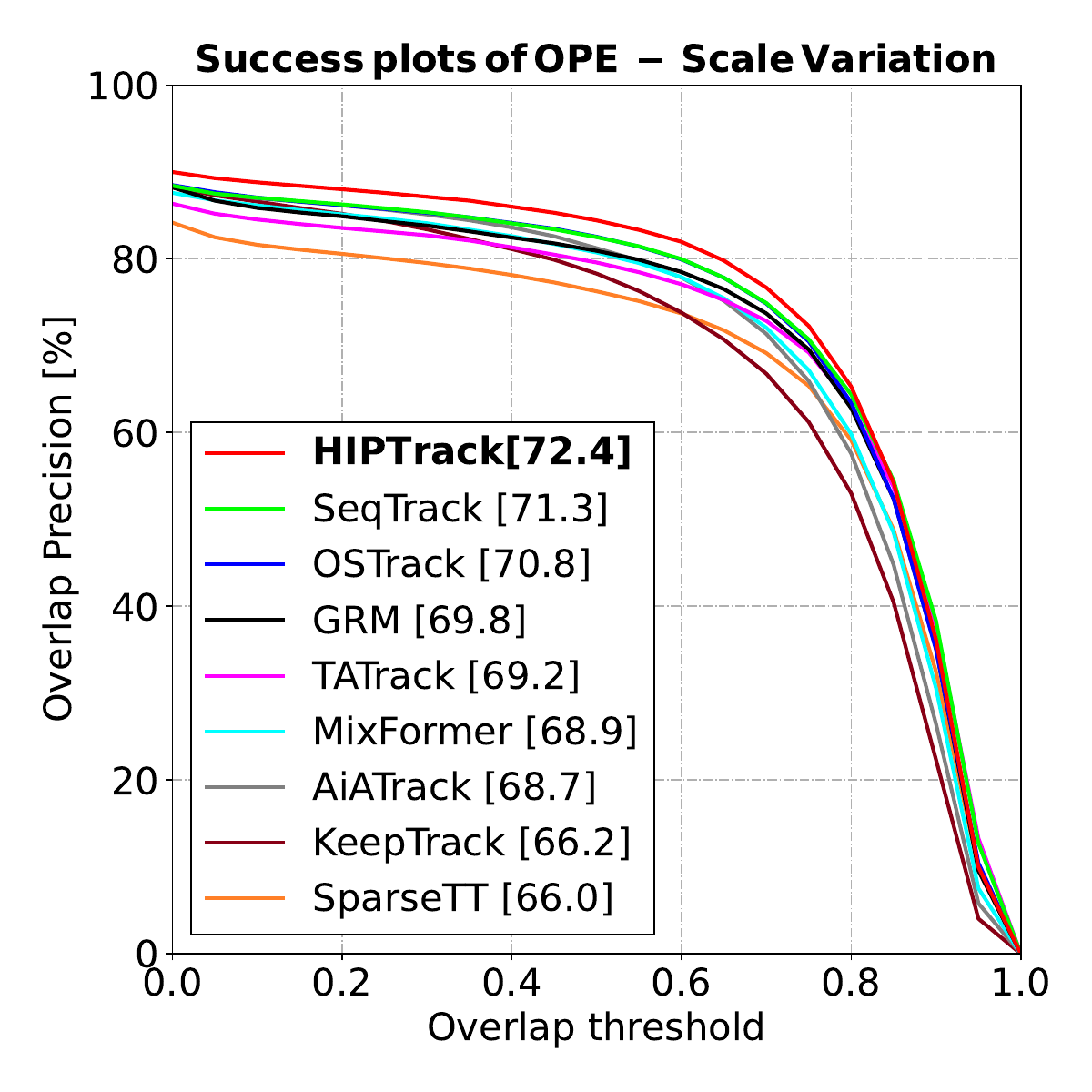}%以这幅图的0.5倍大小输出
\end{minipage}
}
{
\begin{minipage}{4.1cm}
\centering
\includegraphics[scale=0.23]{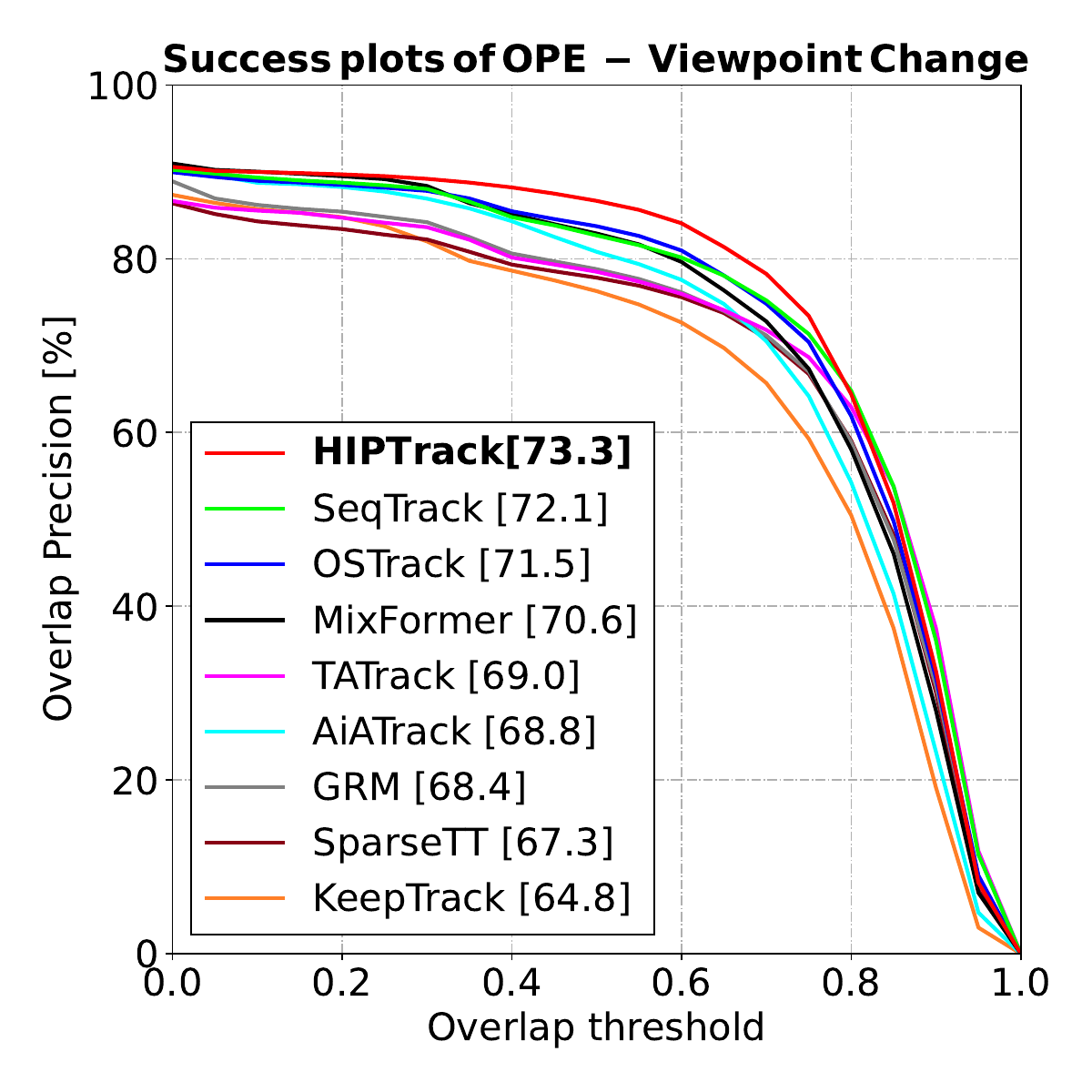}%以这幅图的0.5倍大小输出
\end{minipage}
}
{
\begin{minipage}{4.1cm}
\centering
\includegraphics[scale=0.23]{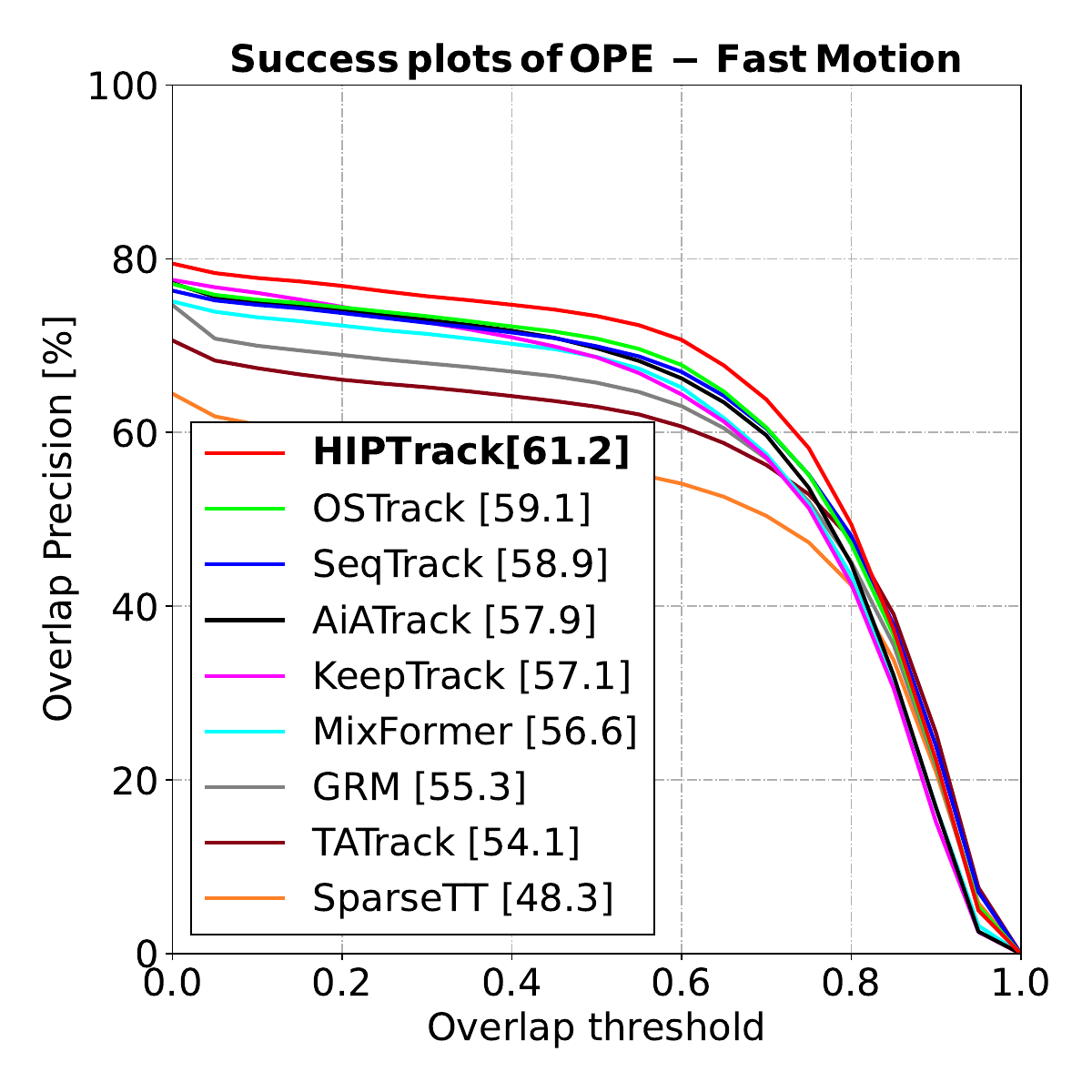}%以这幅图的0.5倍大小输出
\end{minipage}

{
\begin{minipage}{4.1cm}
\centering
\includegraphics[scale=0.23]{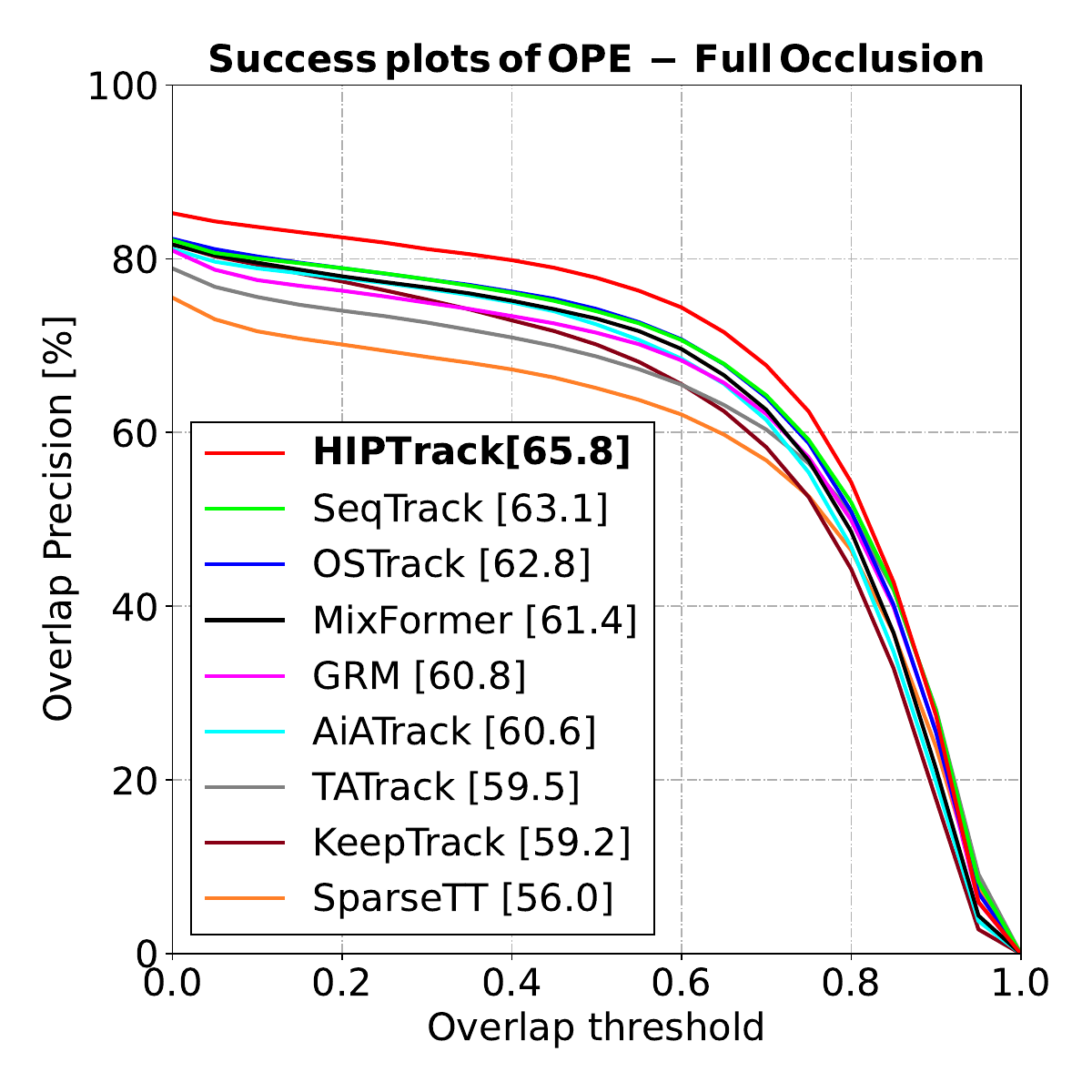}%以这幅图的0.5倍大小输出
\end{minipage}
}
{
\begin{minipage}{4.1cm}
\centering
\includegraphics[scale=0.23]{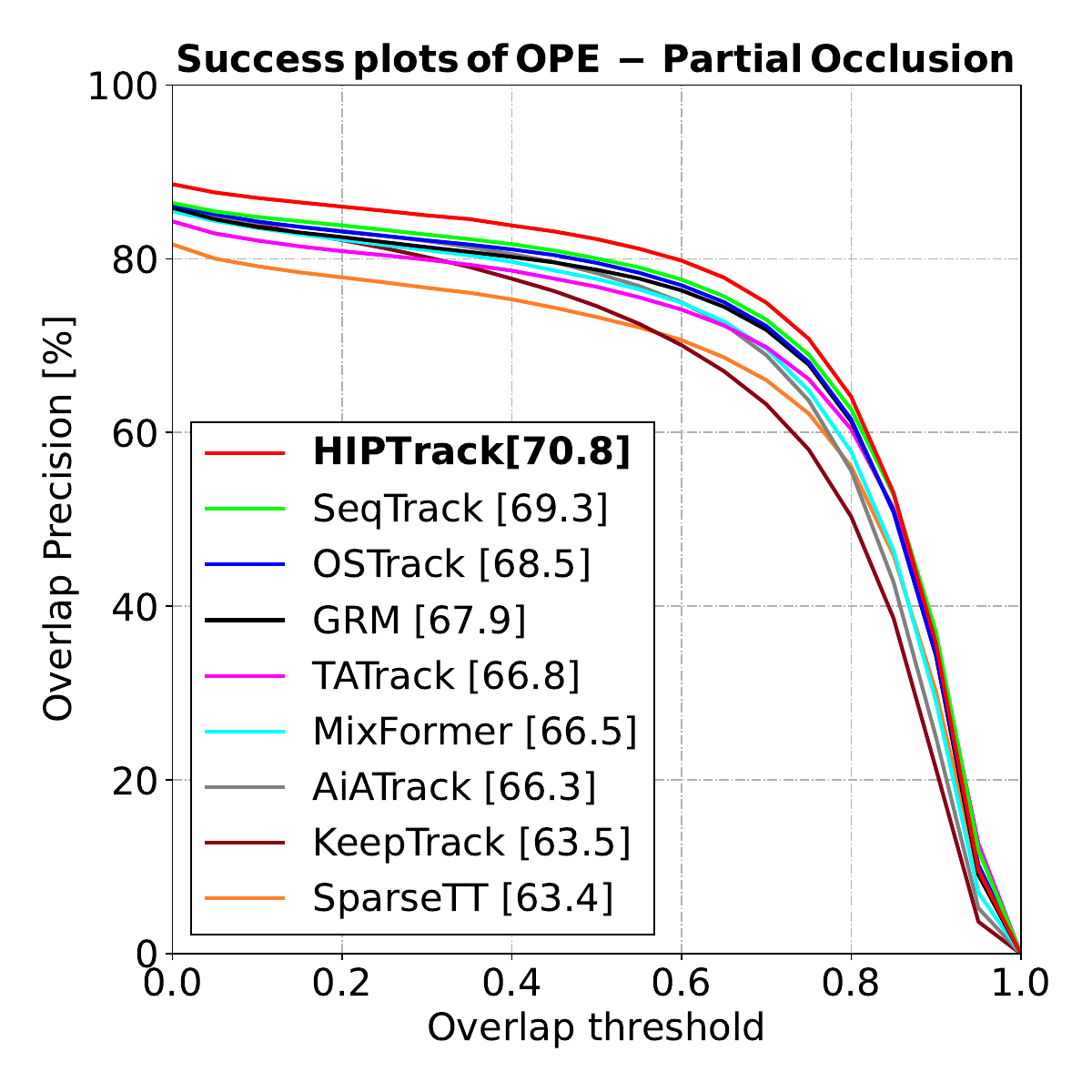}%以这幅图的0.5倍大小输出
\end{minipage}
}
{
\begin{minipage}{4.1cm}
\centering
\includegraphics[scale=0.23]{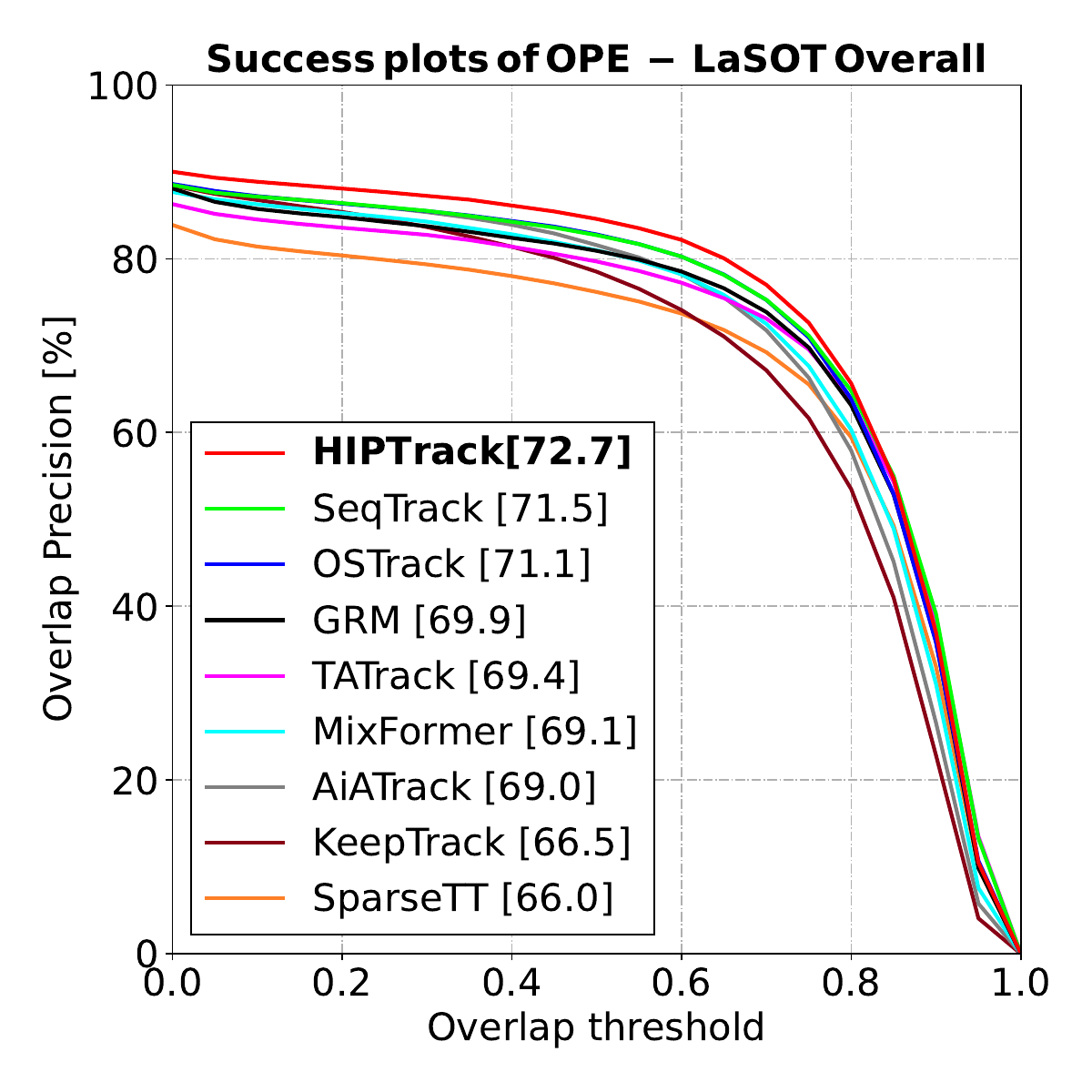}%以这幅图的0.5倍大小输出
\end{minipage}
}
{
\begin{minipage}{4.1cm}
\centering
\includegraphics[scale=0.23]{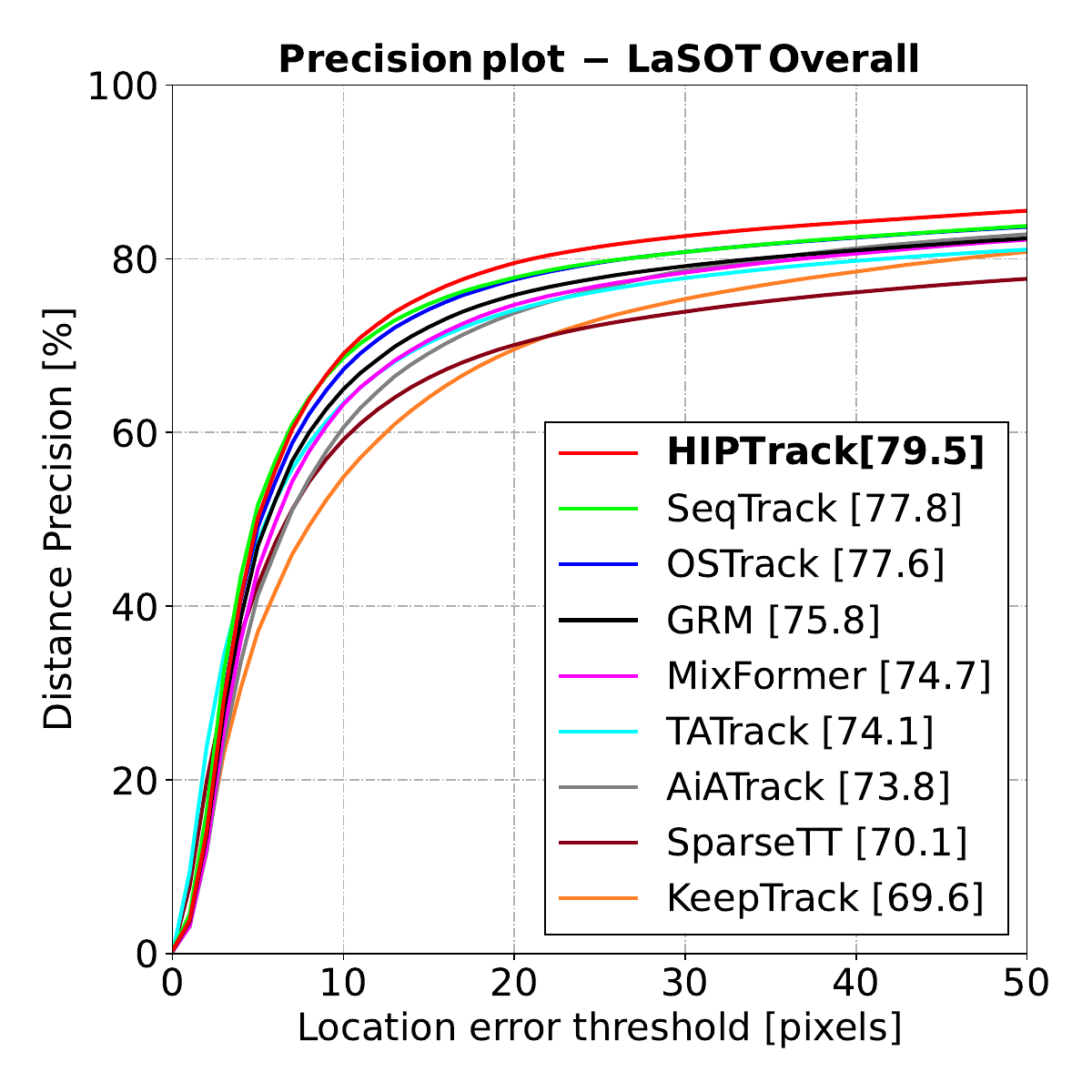}%以这幅图的0.5倍大小输出
\end{minipage}
}
}
\caption{Comparisons of our proposed HIPTrack with other excellent trackers in the success curve on LaSOT \emph{test} split, which includes eleven special scenarios such as Low Resolution, Motion Blur, Scale Variation, etc. We also provide the comparisons of the success and precision curves across the entire LaSOT \emph{test} split.}
\label{fig:comparision of lasot dataset}
\end{figure*}

\subsection{Ablation Studies on Historical Prompt Encoder}

In the historical prompt encoder, we employ a lightweight network $\Phi$ to perform initial encoding on the input 4-channel image tensor and obtain the feature map $\bm{F}$. In the first row of Table \ref{table_other_ablation}, we investigate whether the encoder $\Phi$ can be made even lighter. We replace $\Phi$ from the first three stages of ResNet-18 \cite{he2016deep} with a single convolutional layer. The results in the first and fourth rows indicate that replacing $\Phi$ with a single convolutional layer leads to performance degradation, which suggests that a stronger 
 initial encoder $\Phi$ is required for encoding the historical target features.

After obtaining the feature map $\bm{F_1^{\prime}}$ from the output of the residual block $f_{RB1}$, the historical prompt encoder employs spatial attention and channel attention to enhance the feature map $\bm{F_1^{\prime}}$. In the experiments of the second and third rows in Table \ref{table_other_ablation}, we respectively remove these two branches from the historical prompt encoder to investigate their impact on tracking accuracy. The comparative results of the second, third, and fourth rows indicate that incorporating channel attention and spatial attention both yield positive benefits in tracking accuracy.

%The fusion block we used employs a residual-connected convolutional block for channel dimension reduction and feature fusion. It also incorporates weighted aggregations in both channel and spatial dimensions. To validate the significance of the residual block, channel-wise and spatial-wise weighting, we individually replaced the residual block with a single-layer convolution and removed the channel-wise and spatial-wise weighting. Tab.\ref{table_other_ablation} indicates that all three modules contribute to the improvement in tracking performance.

\begin{table}[!h]\small
    \centering
    \caption{Ablation studies on lightweight encoder $\Phi$ and whether to use channel and spatial attention on LaSOT \emph{test} set.}
    \vspace{-1ex}
    %caption 太长了，精简一下，有哪几个模块就说： Ablation study on ....
    \setlength{\tabcolsep}{1.5mm}
    \begin{tabular}{c|ccc|ccc}
    \Xhline{2pt}
        \textbf{\#} & $\Phi$ & Channel & Spatial & \textbf{AUC}(\%) & $\mathbf{P_{Norm}}$(\%) & $\mathbf{P}$(\%) \\
        \Xhline{1pt}
        \textbf{1} & \ding{56} & \ding{52} & \ding{52} & 72.1 & 82.2 & 78.7\\
        \textbf{2} & \ding{52} & \ding{56} & \ding{52} & 72.3 & 82.4 & 79.1\\
        \textbf{3} & \ding{52} & \ding{52} & \ding{56} & 72.4 & 82.5 &  79.1 \\
        \textbf{4} & \ding{52} & \ding{52} & \ding{52} & \textbf{72.7} & \textbf{82.9} & \textbf{79.5} \\
    \Xhline{2pt}
    \end{tabular}
    \label{table_other_ablation}
    \vspace{-2ex}
\end{table}

\subsection{Ablation Studies on Historical Prompt Decoder}

\begin{figure*}[]
  \centering
    \subfigure[Qualitative results of three methods when the targets undergo large deformations.]{\includegraphics[width=\textwidth]{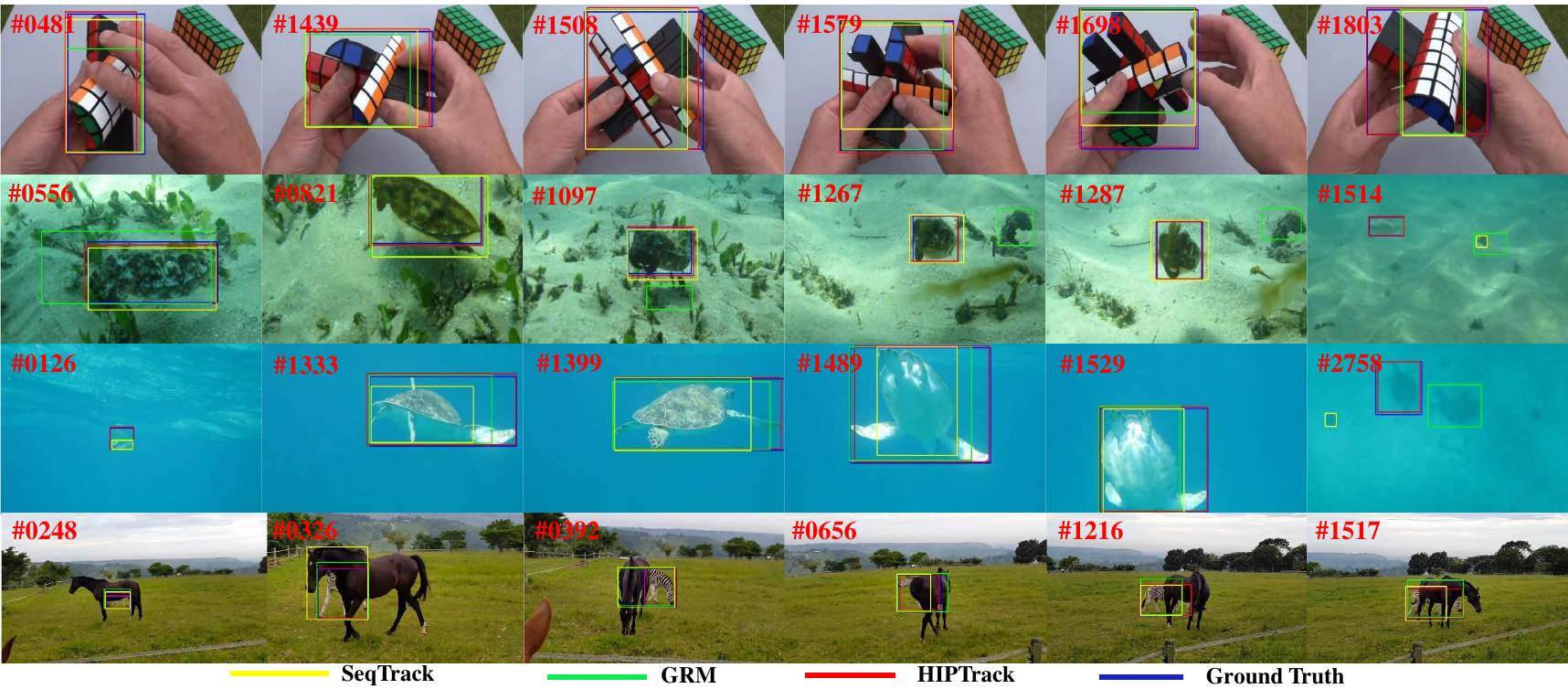}\label{fig-vis-def}} \\
    \subfigure[ Qualitative results of three methods when the targets suffer from partial occlusions.]{\includegraphics[width=\textwidth]{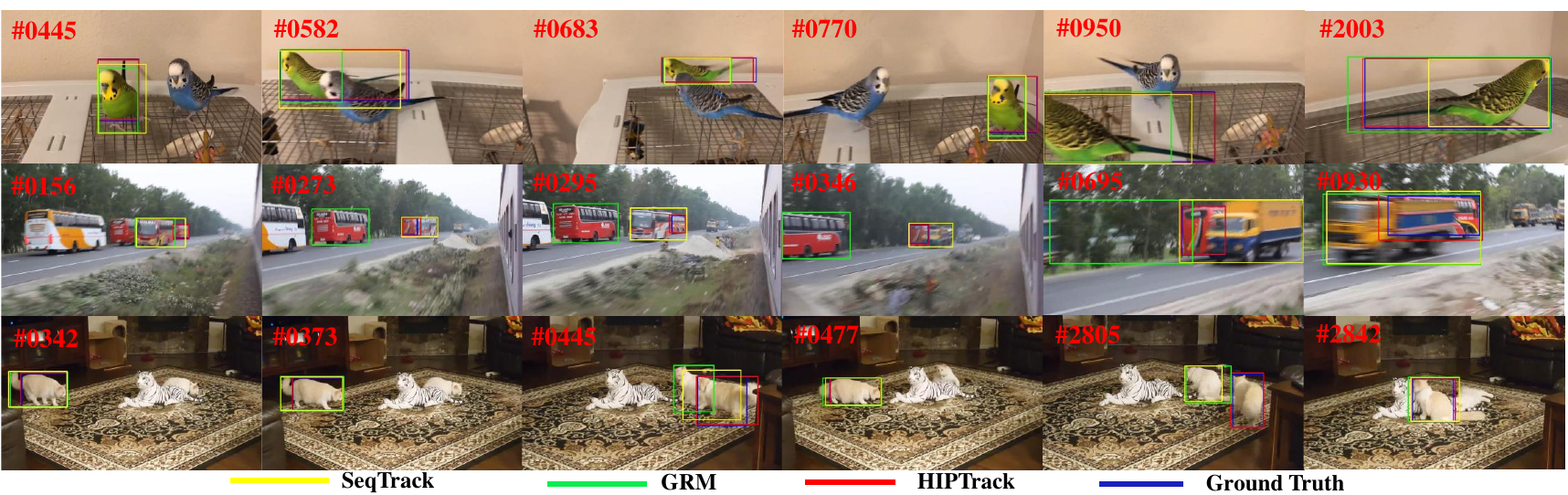}\label{fig-vis-poc}} \\
    \subfigure[ Qualitative results of three methods when the targets have large scale variations.]{\includegraphics[width=\textwidth]{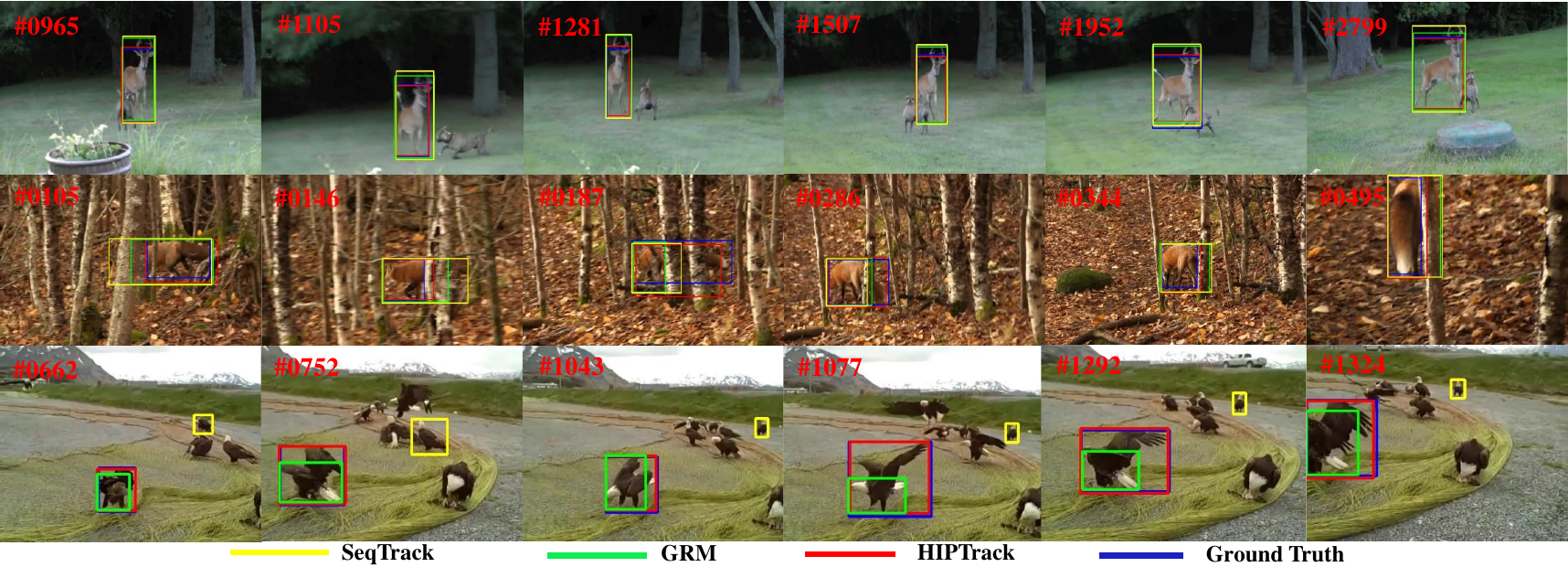}\label{fig-vis-sv}} \\
  \caption{ This figure presents a visual comparison among our method, SeqTrack \cite{Chen_2023_CVPR_seqtrack} and GRM \cite{Gao_2023_CVPR_GRM} in the challenges of target deformation, partial occlusion and scale variation. It demonstrates that our method achieves more effective and accurate tracking in the aforementioned challenging scenarios. Zoom in for better view.}
    \label{fig:quantitative}
    \vspace{0.2in}
\end{figure*}

The historical prompt decoder is utilized to store the historical target features and adaptively aggregate them with the current search region feature to generate the historical prompt. In Table \ref{table-membank}, we investigate the impact of different memory bank sizes on the tracking performance. In Table \ref{ablation_interval}, we investigate the impact of different memory update intervals on the tracking performance.

\textbf{Memory Bank Size.} The first four rows of Table \ref{table-membank} indicate that increasing the memory bank size leads to a performance improvement. In our approach, we set the memory bank size to 150 without carefully tuning, which also implies that there is still potential for performance improvement in our approach. The results from the fourth and fifth rows of Table \ref{table-membank} demonstrate that adding more historical information at the initial stage of tracking leads to a slight performance improvement. This may be because the result in the initial stage of tracking usually has higher accuracy, which facilitates the rediscovery of the target after it is lost.

\textbf{Update Interval.} As shown in Table \ref{ablation_interval}, we conduct experiments using different update intervals on LaSOT. We find that setting the update intervals to 5 or 10 resulted in negligible performance improvement. Additionally, lower update intervals require more frequent calls to the historical prompt encoder,
%And with lower update intervals, the model requires more frequent calls to the historical prompt encoder, 
which can diminish efficiency. On the other hand, longer intervals such as 30 lead to a decline in performance. Therefore, we have chosen an update interval of 20, without carefully tuning as well.
%It is also noteworthy that we did not carefully adjust the update interval parameters, so there is still room for performance improvement in our approach.

\begin{table}[!h]\small
    \centering
    \caption{Ablation study on memory bank sizes and whether to preserve the first 10 memory frames on LaSOT \emph{test} set.}
    \setlength{\tabcolsep}{1mm}
    \begin{tabular}{c|c|c|cccc}
    \Xhline{2pt}
        \multirow{2}{*}{\#} & \textbf{Memory Bank} & \multirow{2}{*}{\textbf{init 10}} & \multirow{2}{*}{\textbf{AUC}(\%)} & \multirow{2}{*}{$\mathbf{P_{Norm}}$(\%)} & \multirow{2}{*}{$\mathbf{P}$(\%)}  \\
         & \textbf{Size} & & & & \\
        \Xhline{1pt}
        \textbf{1} & 70 & \ding{52} & 72.5 & 82.7 & 79.4 \\
        \textbf{2} & 100 & \ding{52} & 72.5 & 82.8 & 79.4\\
        \textbf{3} & 120 & \ding{52} & 72.6 & 82.8 & 79.5\\
        \textbf{4} & 150 & \ding{52} & \textbf{72.7} & \textbf{82.9} & \textbf{79.5}\\
        \textbf{5} & 150 & \ding{56} & 72.7 & 82.8 & 79.5 \\
    \Xhline{2pt}
    \end{tabular}
    \label{table-membank}
\end{table}

\begin{table}[!h]\small
    \centering
    \caption{Ablation studies on different update intervals on LaSOT \emph{test} set.}
    \setlength{\tabcolsep}{4mm}
    \begin{tabular}{c|cccc}
    \Xhline{1.5pt}
        \textbf{Interval} & \textbf{5} & \textbf{10} & \textbf{20} & \textbf{30}  \\
        \Xhline{0.5pt}
        AUC(\%) & 72.7 & 72.7 & \textbf{72.7} & 72.6 \\
        $P_{Norm}$(\%) & 82.9 & 82.9 & \textbf{82.9} & 82.5 \\
        $P$(\%) & 79.5 & 79.5 & \textbf{79.5} & 79.0 \\
    \Xhline{1.5pt}
    \end{tabular}
    \label{ablation_interval}
\end{table}

\section{More Detailed Results in Different Attribute Scenes on LaSOT}\label{sec2}

\begin{figure*}[!ht]
%是可选项 h表示的是here在这里插入，t表示的是在页面的顶部插入
\centering
\includegraphics[scale=0.55]{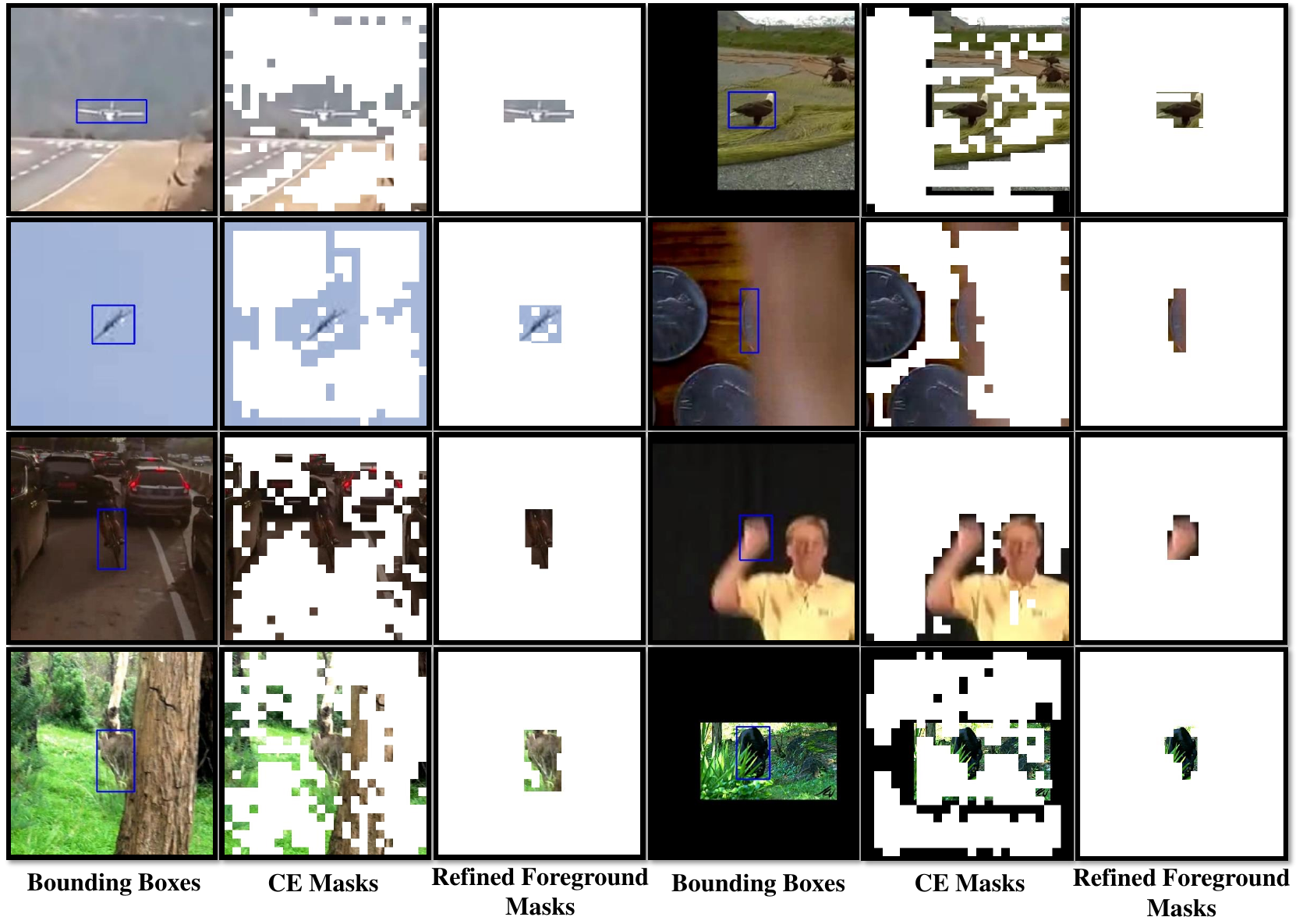}
\caption{ 
Visualization results of refined foreground masks. The construction process of refined foreground masks involves combining the bounding box masks generated from the predicted bounding boxes and the CE masks obtained from the candidate elimination module within the feature extraction network. The combining process is performed using the bitwise $\mathrm{and}$ operation. 
}
\label{fig:masks}
\end{figure*}

\begin{figure*}[!th]
%是可选项 h表示的是here在这里插入，t表示的是在页面的顶部插入
\centering
\includegraphics[scale=0.53]{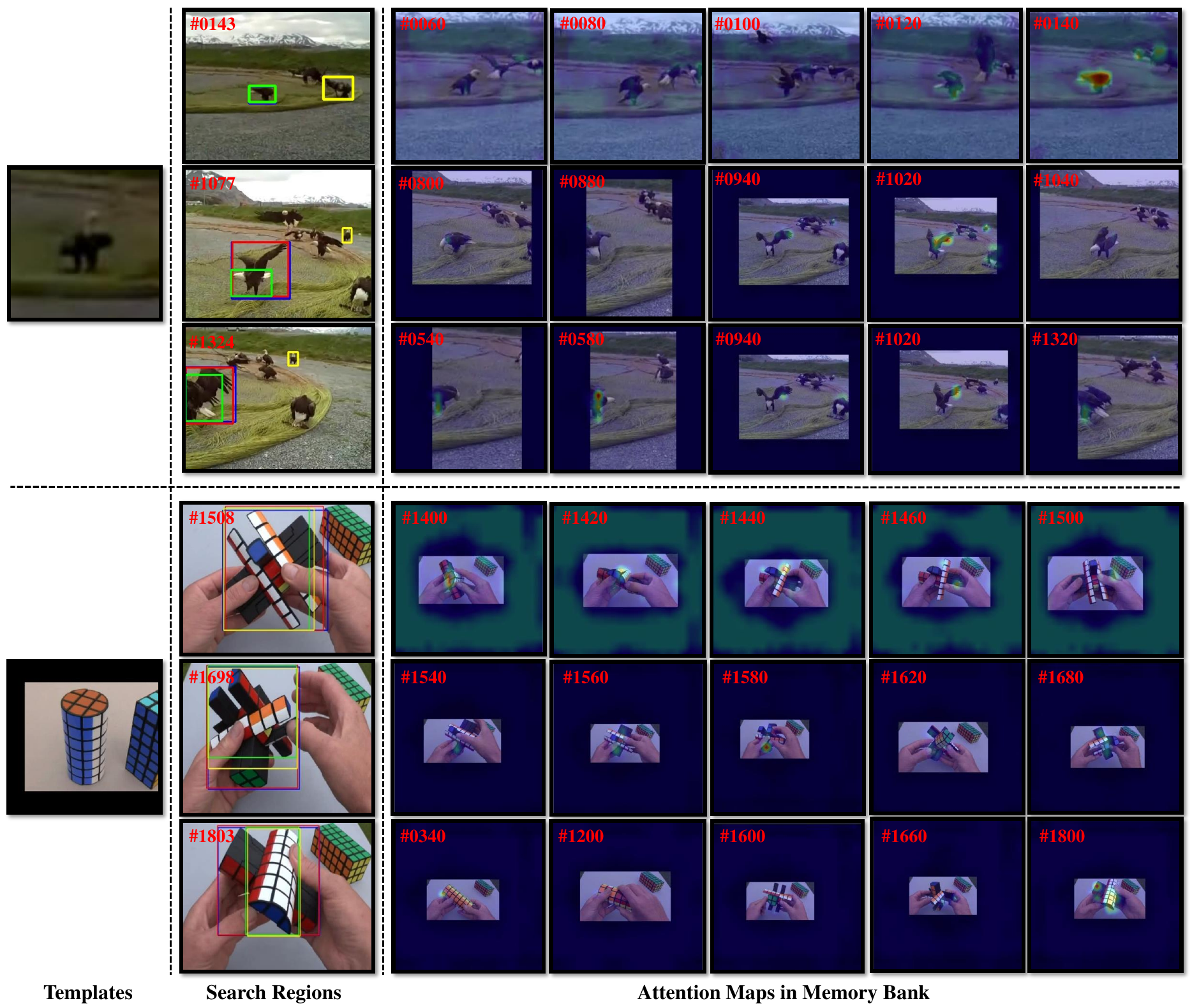}
\caption{ 
Visualization results of memory bank attention maps across different tracking frames in each video. We select the top 5 memory frames with the highest overall attention weights for visualization, arranging them in chronological order. Zoom in for a clearer view.
}
\label{fig:membank}
\end{figure*}

In Figure \ref{fig:comparision of lasot dataset}, we present more detailed quantitative comparisons of the success curves between our proposed HIPTrack and other excellent trackers SeqTrack \cite{Chen_2023_CVPR_seqtrack}, GRM \cite{Gao_2023_CVPR_GRM}, TATrack \cite{he2023target_TATrack}, MixFormer \cite{Cui_2022_MixFormer}, KeepTrack \cite{mayer2021learning}, OSTrack \cite{ye_2022_joint}, AiATrack \cite{gao2022aiatrack}, and SparseTT \cite{fu2022sparsett} across various attribute scenes in LaSOT \cite{fan2019lasot} \emph{test} split. Figure \ref{fig:comparision of lasot dataset} illustrates that our HIPTrack outperforms other trackers across all subsets of videos with special attributes in LaSOT. Particularly, when dealing with scenarios involving partial occlusion, full occlusion, motion blur, and scale variation, our method surpasses the second best method by \textbf{+1.5\%}, \textbf{+2.7\%}, \textbf{+1.4\%}, and \textbf{+1.1\%} AUC, respectively.
The results in Figure \ref{fig:comparision of lasot dataset} demonstrate that our proposed HIPTrack maintains a high level of tracking accuracy and exhibits strong robustness in scenarios involving target appearance variations. 

Furthermore, when the target goes out of view, our proposed HIPTrack also exhibits a performance improvement of \textbf{+1.1\%} AUC compared to the second best method, which means that our proposed HIPTrack has a strong ability to rediscover the lost target.
Figure \ref{fig:comparision of lasot dataset} also includes the comparisons of the success and precision curves between our proposed HIPTrack and other approaches across the entire LaSOT \emph{test} split. Our method achieves the highest performance in both two metrics.

\section{More Qualitative Results}\label{sec3}

\subsection{Tracking Results}

In order to visually highlight the advantages of our method over existing approaches in challenging scenarios, we provide additional visualization results in Figure \ref{fig:quantitative}. All videos are from the \emph{test} split of LaSOT. We compare our proposed HIPTrack with GRM \cite{Gao_2023_CVPR_GRM} and SeqTrack \cite{Chen_2023_CVPR_seqtrack} in terms of performance when the target undergoes deformation, occlusion, and scale variation. All the selected video segments are challenging , as described below:

\begin{itemize}
    \item Figure \ref{fig-vis-def} demonstrates the tracking results of three methods when the target suffers large deformations.
    \item Figure \ref{fig-vis-poc} demonstrates the tracking results of three methods when the target suffers partial occlusions.
    \item Figure \ref{fig-vis-sv} demonstrates the tracking results of three methods when the target suffers large scale variations.
\end{itemize}

%In this work, we have identified that the primary bottleneck limiting the performance improvement of current Transformer-based one-stream trackers is not in their representation capacity but in their inability to effectively introduce abundant and precise historical information. Based on this observation, we propose the Historical Information Prompter (HIP) module. HIP module introduces historical positional information through the construction of refined target masks, and simultaneously incorporates historical visual information. This is achieved by employing a lightweight prompt value encoder to generate historical information prompts. Notably, this entire process introduces only a small number of parameters and requires no training of the backbone network, making it readily applicable to existing trackers. Building upon HIP module, we introduce HIPTrack, achieving state-of-the-art performance on datasets such as LaSOT and GOT-10k.
%We hope that this work will stimulate further research in the domain of visual tracking, particularly in the integration of historical information and model prompting techniques.

%\bigskip
%\noindent Thank you for reading these instructions carefully. We look forward to receiving your electronic files!

\subsection{Refined Foreground Masks}

Our proposed historical prompt encoder utilizes the candidate elimination (CE) module within the feature extraction network to filter out background image patches to construct a CE Mask. The Bounding box mask is created based on the predicted bounding box of the current frame. These two masks are then combined using bitwise $\mathrm{and}$ operation, resulting in a refined target foreground mask. To investigate whether the refined foreground mask accurately captures the position information of the target, we visualize the predicted bounding boxes, CE Masks, and refined foreground masks in Figure \ref{fig:masks}. The visualization results in Figure \ref{fig:masks} demonstrate that the refined foreground mask effectively filters out the majority of background regions, providing a more precise depiction of the position information of the target.

\subsection{Attention Maps in Memory Bank}

In the main body of this paper, we present visualization results of a subset of memory bank attention maps. In Figure \ref{fig:membank}, we further illustrate the visualization results of memory bank attention maps across different tracking frames within the same video.

As shown in Figure \ref{fig:membank}, the first row of results in the first video demonstrates that when the target has not undergone significant deformations or scale changes in recent frames, the most recent memory frame receives noticeably higher attention. However, the second and third rows in the first video indicate that when the target undergoes drastic changes in appearance, the historical prompt decoder directs attention towards earlier historical memory frames, thereby enhancing the prediction accuracy of the tracker. 

In the second video, when the target undergoes severe deformations, the historical prompt decoder adaptively directs attention to the boundary regions of the target within the historical memory frames, leading to a significant improvement in the precision of boundary prediction. A similar phenomenon can also be observed in the second row of the first video.
{
    \small
    \bibliographystyle{ieeenat_fullname}
    \bibliography{main}

\begin{thebibliography}{52}
\providecommand{\natexlab}[1]{#1}
\providecommand{\url}[1]{\texttt{#1}}
\expandafter\ifx\csname urlstyle\endcsname\relax
  \providecommand{\doi}[1]{doi: #1}\else
  \providecommand{\doi}{doi: \begingroup \urlstyle{rm}\Url}\fi

\bibitem[Bertinetto et~al.(2016)Bertinetto, Valmadre, Henriques, Vedaldi, and Torr]{bertinetto2016fully}
Luca Bertinetto, Jack Valmadre, Joao~F Henriques, Andrea Vedaldi, and Philip~HS Torr.
\newblock Fully-convolutional siamese networks for object tracking.
\newblock In \emph{ECCV}, pages 850--865, 2016.

\bibitem[Bhat et~al.(2019)Bhat, Danelljan, Gool, and Timofte]{bhat2019learning}
Goutam Bhat, Martin Danelljan, Luc~Van Gool, and Radu Timofte.
\newblock Learning discriminative model prediction for tracking.
\newblock In \emph{ICCV}, pages 6182--6191, 2019.

\bibitem[Cai et~al.(2023)Cai, Liu, Tang, and Wu]{Cai_2023_ICCV_ROMTrack}
Yidong Cai, Jie Liu, Jie Tang, and Gangshan Wu.
\newblock Robust object modeling for visual tracking.
\newblock In \emph{Proceedings of the IEEE/CVF International Conference on Computer Vision (ICCV)}, pages 9589--9600, 2023.

\bibitem[Chen et~al.(2022)Chen, Li, Bai, Qiao, Shen, Li, Gan, Wu, and Ouyang]{chen2022backbone}
Boyu Chen, Peixia Li, Lei Bai, Lei Qiao, Qiuhong Shen, Bo Li, Weihao Gan, Wei Wu, and Wanli Ouyang.
\newblock Backbone is all your need: a simplified architecture for visual object tracking.
\newblock In \emph{Computer Vision--ECCV 2022: 17th European Conference, Tel Aviv, Israel, October 23--27, 2022, Proceedings, Part XXII}, pages 375--392. Springer, 2022.

\bibitem[Chen et~al.(2021)Chen, Yan, Zhu, Wang, Yang, and Lu]{chen2021transformer}
Xin Chen, Bin Yan, Jiawen Zhu, Dong Wang, Xiaoyun Yang, and Huchuan Lu.
\newblock Transformer tracking.
\newblock In \emph{CVPR}, pages 8126--8135, 2021.

\bibitem[Chen et~al.(2023)Chen, Peng, Wang, Lu, and Hu]{Chen_2023_CVPR_seqtrack}
Xin Chen, Houwen Peng, Dong Wang, Huchuan Lu, and Han Hu.
\newblock Seqtrack: Sequence to sequence learning for visual object tracking.
\newblock In \emph{Proceedings of the IEEE/CVF Conference on Computer Vision and Pattern Recognition (CVPR)}, pages 14572--14581, 2023.

\bibitem[Cheng et~al.(2021)Cheng, Tai, and Tang]{cheng2021rethinking}
Ho~Kei Cheng, Yu-Wing Tai, and Chi-Keung Tang.
\newblock Rethinking space-time networks with improved memory coverage for efficient video object segmentation.
\newblock \emph{Advances in Neural Information Processing Systems}, 34:\penalty0 11781--11794, 2021.

\bibitem[Cui et~al.(2022)Cui, Jiang, Wang, and Wu]{Cui_2022_MixFormer}
Yutao Cui, Cheng Jiang, Limin Wang, and Gangshan Wu.
\newblock Mixformer: End-to-end tracking with iterative mixed attention.
\newblock In \emph{Proceedings of the IEEE/CVF Conference on Computer Vision and Pattern Recognition (CVPR)}, pages 13608--13618, 2022.

\bibitem[Dai et~al.(2020)Dai, Zhang, Wang, Li, Lu, and Yang]{dai2020high}
Kenan Dai, Yunhua Zhang, Dong Wang, Jianhua Li, Huchuan Lu, and Xiaoyun Yang.
\newblock High-performance long-term tracking with meta-updater.
\newblock In \emph{CVPR}, pages 6298--6307, 2020.

\bibitem[Dosovitskiy et~al.(2021)Dosovitskiy, Beyer, Kolesnikov, Weissenborn, Zhai, Unterthiner, Dehghani, Minderer, Heigold, Gelly, et~al.]{dosovitskiy2020image}
Alexey Dosovitskiy, Lucas Beyer, Alexander Kolesnikov, Dirk Weissenborn, Xiaohua Zhai, Thomas Unterthiner, Mostafa Dehghani, Matthias Minderer, Georg Heigold, Sylvain Gelly, et~al.
\newblock An image is worth 16x16 words: Transformers for image recognition at scale.
\newblock In \emph{ICLR}, 2021.

\bibitem[Fan et~al.(2019)Fan, Lin, Yang, Chu, Deng, Yu, Bai, Xu, Liao, and Ling]{fan2019lasot}
Heng Fan, Liting Lin, Fan Yang, Peng Chu, Ge Deng, Sijia Yu, Hexin Bai, Yong Xu, Chunyuan Liao, and Haibin Ling.
\newblock Lasot: A high-quality benchmark for large-scale single object tracking.
\newblock In \emph{CVPR}, pages 5374--5383, 2019.

\bibitem[Fan et~al.(2021)Fan, Bai, Lin, Yang, Chu, Deng, Yu, Huang, Liu, Xu, et~al.]{fan2021lasot}
Heng Fan, Hexin Bai, Liting Lin, Fan Yang, Peng Chu, Ge Deng, Sijia Yu, Mingzhen Huang, Juehuan Liu, Yong Xu, et~al.
\newblock Lasot: A high-quality large-scale single object tracking benchmark.
\newblock \emph{International Journal of Computer Vision}, 129:\penalty0 439--461, 2021.

\bibitem[Fu et~al.(2021)Fu, Liu, Fu, and Wang]{fu2021stmtrack}
Zhihong Fu, Qingjie Liu, Zehua Fu, and Yunhong Wang.
\newblock Stmtrack: Template-free visual tracking with space-time memory networks.
\newblock In \emph{CVPR}, pages 13774--13783, 2021.

\bibitem[Fu et~al.(2022)Fu, Fu, Liu, Cai, and Wang]{fu2022sparsett}
Zhihong Fu, Zehua Fu, Qingjie Liu, Wenrui Cai, and Yunhong Wang.
\newblock Sparsett: Visual tracking with sparse transformers.
\newblock In \emph{Proceedings of the Thirtieth International Joint Conference on Artificial Intelligence, {IJCAI-22}}, pages 905--912, 2022.

\bibitem[Gao et~al.(2022)Gao, Zhou, Ma, Wang, and Yuan]{gao2022aiatrack}
Shenyuan Gao, Chunluan Zhou, Chao Ma, Xinggang Wang, and Junsong Yuan.
\newblock Aiatrack: Attention in attention for transformer visual tracking.
\newblock In \emph{Computer Vision--ECCV 2022: 17th European Conference, Tel Aviv, Israel, October 23--27, 2022, Proceedings, Part XXII}, pages 146--164. Springer, 2022.

\bibitem[Gao et~al.(2023)Gao, Zhou, and Zhang]{Gao_2023_CVPR_GRM}
Shenyuan Gao, Chunluan Zhou, and Jun Zhang.
\newblock Generalized relation modeling for transformer tracking.
\newblock In \emph{Proceedings of the IEEE/CVF Conference on Computer Vision and Pattern Recognition (CVPR)}, pages 18686--18695, 2023.

\bibitem[He et~al.(2016)He, Zhang, Ren, and Sun]{he2016deep}
Kaiming He, Xiangyu Zhang, Shaoqing Ren, and Jian Sun.
\newblock Deep residual learning for image recognition.
\newblock In \emph{CVPR}, pages 770--778, 2016.

\bibitem[He et~al.(2023)He, Zhang, Xie, Li, and Wang]{he2023target_TATrack}
Kaijie He, Canlong Zhang, Sheng Xie, Zhixin Li, and Zhiwen Wang.
\newblock Target-aware tracking with long-term context attention.
\newblock 2023.

\bibitem[Huang et~al.(2019)Huang, Zhao, and Huang]{huang2019got}
Lianghua Huang, Xin Zhao, and Kaiqi Huang.
\newblock Got-10k: A large high-diversity benchmark for generic object tracking in the wild.
\newblock \emph{TPAMI}, 2019.

\bibitem[Jia et~al.(2022)Jia, Tang, Chen, Cardie, Belongie, Hariharan, and Lim]{jia2022visualPromptTuning}
Menglin Jia, Luming Tang, Bor-Chun Chen, Claire Cardie, Serge Belongie, Bharath Hariharan, and Ser-Nam Lim.
\newblock Visual prompt tuning.
\newblock In \emph{European Conference on Computer Vision}, pages 709--727. Springer, 2022.

\bibitem[Khattak et~al.(2023)Khattak, Rasheed, Maaz, Khan, and Khan]{Khattak_2023_CVPR_MaPLe}
Muhammad~Uzair Khattak, Hanoona Rasheed, Muhammad Maaz, Salman Khan, and Fahad~Shahbaz Khan.
\newblock Maple: Multi-modal prompt learning.
\newblock In \emph{Proceedings of the IEEE/CVF Conference on Computer Vision and Pattern Recognition (CVPR)}, pages 19113--19122, 2023.

\bibitem[Kiani~Galoogahi et~al.(2017)Kiani~Galoogahi, Fagg, Huang, Ramanan, and Lucey]{Galoogahi_2017_ICCV_nfs}
Hamed Kiani~Galoogahi, Ashton Fagg, Chen Huang, Deva Ramanan, and Simon Lucey.
\newblock Need for speed: A benchmark for higher frame rate object tracking.
\newblock In \emph{Proceedings of the IEEE International Conference on Computer Vision (ICCV)}, 2017.

\bibitem[Lester et~al.(2021)Lester, Al-Rfou, and Constant]{lester2021power}
Brian Lester, Rami Al-Rfou, and Noah Constant.
\newblock The power of scale for parameter-efficient prompt tuning.
\newblock \emph{arXiv preprint arXiv:2104.08691}, 2021.

\bibitem[Li et~al.(2019)Li, Wu, Wang, Zhang, Xing, and Yan]{li2019siamrpn++}
Bo Li, Wei Wu, Qiang Wang, Fangyi Zhang, Junliang Xing, and Junjie Yan.
\newblock Siamrpn++: Evolution of siamese visual tracking with very deep networks.
\newblock In \emph{CVPR}, pages 4282--4291, 2019.

\bibitem[Li and Liang(2021)]{li2021prefix}
Xiang~Lisa Li and Percy Liang.
\newblock Prefix-tuning: Optimizing continuous prompts for generation.
\newblock In \emph{Proceedings of the 59th Annual Meeting of the Association for Computational Linguistics and the 11th International Joint Conference on Natural Language Processing (Volume 1: Long Papers)}, pages 4582--4597, 2021.

\bibitem[Lin et~al.(2022)Lin, Fan, Zhang, Xu, and Ling]{lin2022swintrack}
Liting Lin, Heng Fan, Zhipeng Zhang, Yong Xu, and Haibin Ling.
\newblock Swintrack: A simple and strong baseline for transformer tracking.
\newblock \emph{Advances in Neural Information Processing Systems}, 35:\penalty0 16743--16754, 2022.

\bibitem[Lin et~al.(2014)Lin, Maire, Belongie, Hays, Perona, Ramanan, Doll{\'a}r, and Zitnick]{lin2014microsoft}
Tsung-Yi Lin, Michael Maire, Serge Belongie, James Hays, Pietro Perona, Deva Ramanan, Piotr Doll{\'a}r, and C~Lawrence Zitnick.
\newblock Microsoft coco: Common objects in context.
\newblock In \emph{ECCV}, pages 740--755, 2014.

\bibitem[Lin et~al.(2017)Lin, Goyal, Girshick, He, and Doll{\'a}r]{lin2017focal}
Tsung-Yi Lin, Priya Goyal, Ross Girshick, Kaiming He, and Piotr Doll{\'a}r.
\newblock Focal loss for dense object detection.
\newblock In \emph{ICCV}, pages 2980--2988, 2017.

\bibitem[Mayer et~al.(2021)Mayer, Danelljan, Paudel, and Van~Gool]{mayer2021learning}
Christoph Mayer, Martin Danelljan, Danda~Pani Paudel, and Luc Van~Gool.
\newblock Learning target candidate association to keep track of what not to track.
\newblock In \emph{Proceedings of the IEEE/CVF International Conference on Computer Vision}, pages 13444--13454, 2021.

\bibitem[Mayer et~al.(2022)Mayer, Danelljan, Bhat, Paul, Paudel, Yu, and Van~Gool]{mayer2022transforming}
Christoph Mayer, Martin Danelljan, Goutam Bhat, Matthieu Paul, Danda~Pani Paudel, Fisher Yu, and Luc Van~Gool.
\newblock Transforming model prediction for tracking.
\newblock In \emph{Proceedings of the IEEE/CVF conference on computer vision and pattern recognition}, pages 8731--8740, 2022.

\bibitem[Mueller et~al.(2016)Mueller, Smith, and Ghanem]{mueller2016benchmark}
Matthias Mueller, Neil Smith, and Bernard Ghanem.
\newblock A benchmark and simulator for uav tracking.
\newblock In \emph{ECCV}, pages 445--461, 2016.

\bibitem[Muller et~al.(2018)Muller, Bibi, Giancola, Alsubaihi, and Ghanem]{muller2018trackingnet}
Matthias Muller, Adel Bibi, Silvio Giancola, Salman Alsubaihi, and Bernard Ghanem.
\newblock Trackingnet: A large-scale dataset and benchmark for object tracking in the wild.
\newblock In \emph{ECCV}, pages 300--317, 2018.

\bibitem[Oh et~al.(2019)Oh, Lee, Xu, and Kim]{oh2019video}
Seoung~Wug Oh, Joon-Young Lee, Ning Xu, and Seon~Joo Kim.
\newblock Video object segmentation using space-time memory networks.
\newblock In \emph{ICCV}, pages 9226--9235, 2019.

\bibitem[Radford et~al.(2021)Radford, Kim, Hallacy, Ramesh, Goh, Agarwal, Sastry, Askell, Mishkin, Clark, Krueger, and Sutskever]{pmlr-v139-radford21a}
Alec Radford, Jong~Wook Kim, Chris Hallacy, Aditya Ramesh, Gabriel Goh, Sandhini Agarwal, Girish Sastry, Amanda Askell, Pamela Mishkin, Jack Clark, Gretchen Krueger, and Ilya Sutskever.
\newblock Learning transferable visual models from natural language supervision.
\newblock In \emph{Proceedings of the 38th International Conference on Machine Learning}, pages 8748--8763. PMLR, 2021.

\bibitem[Rao et~al.(2022)Rao, Zhao, Chen, Tang, Zhu, Huang, Zhou, and Lu]{Rao_2022_CVPR_DenseCLIP}
Yongming Rao, Wenliang Zhao, Guangyi Chen, Yansong Tang, Zheng Zhu, Guan Huang, Jie Zhou, and Jiwen Lu.
\newblock Denseclip: Language-guided dense prediction with context-aware prompting.
\newblock In \emph{Proceedings of the IEEE/CVF Conference on Computer Vision and Pattern Recognition (CVPR)}, pages 18082--18091, 2022.

\bibitem[Rezatofighi et~al.(2019)Rezatofighi, Tsoi, Gwak, Sadeghian, Reid, and Savarese]{rezatofighi2019generalized}
Hamid Rezatofighi, Nathan Tsoi, JunYoung Gwak, Amir Sadeghian, Ian Reid, and Silvio Savarese.
\newblock Generalized intersection over union: A metric and a loss for bounding box regression.
\newblock In \emph{CVPR}, pages 658--666, 2019.

\bibitem[Song et~al.(2023)Song, Luo, Yu, Chen, and Yang]{Song_2023_AAAI_CTTrack}
Zikai Song, Run Luo, Junqing Yu, Yi-Ping~Phoebe Chen, and Wei Yang.
\newblock Compact transformer tracker with correlative masked modeling.
\newblock In \emph{Proceedings of the AAAI Conference on Artificial Intelligence (AAAI)}, 2023.

\bibitem[Wang et~al.(2021)Wang, Zhou, Wang, and Li]{wang2021transformer}
Ning Wang, Wengang Zhou, Jie Wang, and Houqiang Li.
\newblock Transformer meets tracker: Exploiting temporal context for robust visual tracking.
\newblock In \emph{CVPR}, pages 1571--1580, 2021.

\bibitem[Wang et~al.(2022)Wang, Zhang, Lee, Zhang, Sun, Ren, Su, Perot, Dy, and Pfister]{wang2022learningtopromptforcontinual}
Zifeng Wang, Zizhao Zhang, Chen-Yu Lee, Han Zhang, Ruoxi Sun, Xiaoqi Ren, Guolong Su, Vincent Perot, Jennifer Dy, and Tomas Pfister.
\newblock Learning to prompt for continual learning.
\newblock In \emph{Proceedings of the IEEE/CVF Conference on Computer Vision and Pattern Recognition}, pages 139--149, 2022.

\bibitem[Wei et~al.(2023)Wei, Bai, Zheng, Shi, and Gong]{Wei_2023_CVPR_autoregressive}
Xing Wei, Yifan Bai, Yongchao Zheng, Dahu Shi, and Yihong Gong.
\newblock Autoregressive visual tracking.
\newblock In \emph{Proceedings of the IEEE/CVF Conference on Computer Vision and Pattern Recognition (CVPR)}, pages 9697--9706, 2023.

\bibitem[Woo et~al.(2018)Woo, Park, Lee, and Kweon]{Woo_2018_ECCV_CBAM}
Sanghyun Woo, Jongchan Park, Joon-Young Lee, and In~So Kweon.
\newblock Cbam: Convolutional block attention module.
\newblock In \emph{Proceedings of the European Conference on Computer Vision (ECCV)}, 2018.

\bibitem[Wu et~al.(2023)Wu, Yang, Liu, Wu, Shan, and Chan]{Wu_2023_CVPR_dropmae}
Qiangqiang Wu, Tianyu Yang, Ziquan Liu, Baoyuan Wu, Ying Shan, and Antoni~B. Chan.
\newblock Dropmae: Masked autoencoders with spatial-attention dropout for tracking tasks.
\newblock In \emph{Proceedings of the IEEE/CVF Conference on Computer Vision and Pattern Recognition (CVPR)}, pages 14561--14571, 2023.

\bibitem[{Wu} et~al.(2015){Wu}, {Lim}, and {Yang}]{otb2015}
Y. {Wu}, J. {Lim}, and M. {Yang}.
\newblock Object tracking benchmark.
\newblock \emph{TPAMI}, 37\penalty0 (9):\penalty0 1834--1848, 2015.

\bibitem[Xie et~al.(2022)Xie, Wang, Wang, Cao, Yang, and Zeng]{Xie_2022_Correlation}
Fei Xie, Chunyu Wang, Guangting Wang, Yue Cao, Wankou Yang, and Wenjun Zeng.
\newblock Correlation-aware deep tracking.
\newblock In \emph{Proceedings of the IEEE/CVF Conference on Computer Vision and Pattern Recognition (CVPR)}, pages 8751--8760, 2022.

\bibitem[Xu et~al.(2019)Xu, Feng, Wu, and Kittler]{xu2019joint}
Tianyang Xu, Zhen-Hua Feng, Xiao-Jun Wu, and Josef Kittler.
\newblock Joint group feature selection and discriminative filter learning for robust visual object tracking.
\newblock In \emph{ICCV}, pages 7950--7960, 2019.

\bibitem[Xu et~al.(2020)Xu, Wang, Li, Yuan, and Yu]{xu2020siamfc++}
Yinda Xu, Zeyu Wang, Zuoxin Li, Ye Yuan, and Gang Yu.
\newblock Siamfc++: Towards robust and accurate visual tracking with target estimation guidelines.
\newblock In \emph{AAAI}, pages 12549--12556, 2020.

\bibitem[Yan et~al.(2021)Yan, Peng, Fu, Wang, and Lu]{yan2021learning}
Bin Yan, Houwen Peng, Jianlong Fu, Dong Wang, and Huchuan Lu.
\newblock Learning spatio-temporal transformer for visual tracking.
\newblock In \emph{Proceedings of the IEEE/CVF international conference on computer vision}, pages 10448--10457, 2021.

\bibitem[Ye et~al.(2022)Ye, Chang, Ma, Shan, and Chen]{ye_2022_joint}
Botao Ye, Hong Chang, Bingpeng Ma, Shiguang Shan, and Xilin Chen.
\newblock Joint feature learning and relation modeling for tracking: A one-stream framework.
\newblock In \emph{European Conference on Computer Vision}, pages 341--357. Springer, 2022.

\bibitem[Yu et~al.(2021)Yu, Tang, Zheng, Zhu, Wang, Feng, Feng, and Lu]{yu2021high}
Bin Yu, Ming Tang, Linyu Zheng, Guibo Zhu, Jinqiao Wang, Hao Feng, Xuetao Feng, and Hanqing Lu.
\newblock High-performance discriminative tracking with transformers.
\newblock In \emph{ICCV}, pages 9856--9865, 2021.

\bibitem[Zhang et~al.(2020)Zhang, Peng, Fu, Li, and Hu]{zhang2020ocean}
Zhipeng Zhang, Houwen Peng, Jianlong Fu, Bing Li, and Weiming Hu.
\newblock Ocean: Object-aware anchor-free tracking.
\newblock In \emph{ECCV}, 2020.

\bibitem[Zhou et~al.(2022)Zhou, Yang, Loy, and Liu]{zhou2022learning_to_prompt}
Kaiyang Zhou, Jingkang Yang, Chen~Change Loy, and Ziwei Liu.
\newblock Learning to prompt for vision-language models.
\newblock \emph{International Journal of Computer Vision}, 130\penalty0 (9):\penalty0 2337--2348, 2022.

\bibitem[Zhu et~al.(2023)Zhu, Lai, Chen, Wang, and Lu]{Zhu_2023_CVPR_visual_prompt_multimodal_tracking}
Jiawen Zhu, Simiao Lai, Xin Chen, Dong Wang, and Huchuan Lu.
\newblock Visual prompt multi-modal tracking.
\newblock In \emph{Proceedings of the IEEE/CVF Conference on Computer Vision and Pattern Recognition (CVPR)}, pages 9516--9526, 2023.

\end{thebibliography}
}
\end{document}